\documentclass[10pt,twocolumn,letterpaper]{article}
\usepackage{iccv}

\usepackage{graphicx}
\usepackage{amsmath}
\usepackage{amssymb}
\usepackage{booktabs}
\usepackage{multirow}
\usepackage[dvipsnames]{xcolor}
\usepackage{bold-extra}
\usepackage{listings}
\usepackage{placeins}
\usepackage[algoruled]{algorithm2e}
\usepackage{algpseudocode}
\usepackage{xspace}
\usepackage{listings}
\usepackage[accsupp]{axessibility}
\usepackage{subcaption}

\usepackage{caption}
\usepackage[pagebackref,breaklinks,colorlinks]{hyperref}
\hypersetup{
    colorlinks=true,
    citecolor=blue,  
}

\usepackage[capitalize]{cleveref}
\crefname{section}{Sec.}{Secs.}
\Crefname{section}{Section}{Sections}
\Crefname{table}{Table}{Tables}
\crefname{table}{Tab.}{Tabs.}

\iccvfinalcopy 

\def\iccvPaperID{10972} 

\ificcvfinal\fi

\begin{document}

\newif\ifcomments
\commentstrue 
\ifcomments
    \newcommand\todo[1]{\textcolor{red}{[TODO: #1]}}
    \newcommand\irena[1]{\textcolor{blue}{[Irena: #1]}}
    \newcommand\marco[1]{\textcolor{purple}{[Marco: #1]}}
    \newcommand\gabriel[1]{\textcolor{green}{[Gabriel: #1]}}
    \newcommand\scott[1]{\textcolor{orange}{[Scott: #1]}}
\else
    \newcommand\todo[1]{}
    \newcommand\irena[1]{}
    \newcommand\marco[1]{}
    \newcommand\gabriel[1]{}
    \newcommand\scott[1]{}
\fi

\newcommand{\tightparagraph}[1]{\vspace{0.1in}\noindent\textbf{#1}}

\newcommand{\tool}{{\sc AdaVision}\xspace}
\newcommand{\baseline}{{\sc NonAdaptive}\xspace}
\newcommand{\retrieval}{{\sc CLIP-Retrieval}\xspace}
\newcommand{\domino}{{\sc Domino}\xspace}
\newcommand{\dominobert}{\domino~({\sc BERT})\xspace}
\newcommand{\dominoofa}{\domino~({\sc OFA})\xspace}
\newcommand{\laion}{{LAION-5B}\xspace}
\newcommand{\clip}{{CLIP}\xspace}
\newcommand{\gpt}{{GPT-3}\xspace}
\newcommand{\clipvit}{{CLIP ViT-L/14}\xspace}
\newcommand{\vit}{ViT-H/14\xspace}
\newcommand{\resnet}{ResNet-50\xspace}
\newcommand{\ofa}{OFA-Huge\xspace}

\newcommand{\slice}{coherent group}
\newcommand{\slices}{coherent groups}
\newcommand{\Slices}{Coherent groups}
\newcommand{\subpop}{subset}
\newcommand{\subpops}{subsets}
\newcommand{\Subpops}{Subsets}

\newcommand{\inl}{test generation loop\xspace}
\newcommand{\Inl}{Test generation loop\xspace}
\newcommand{\outl}{topic generation loop\xspace}
\newcommand{\Outl}{Topic generation loop\xspace}

\definecolor{light-gray}{gray}{0.925}
\newcommand{\topic}[1]{\setlength\fboxsep{0pt}\colorbox{light-gray}{\lstinline{#1}}}
\newcommand{\y}[1]{\textit{y}}

\newcommand{\class}{{classification}\xspace} 
\newcommand{\detc}{{object detection}\xspace}
\newcommand{\capt}{{image captioning}\xspace}

\newcommand{\sixoverlap}{\{banana, broom, candle, lemon, sandal, wine bottle\}\xspace}

\newcommand{\model}{m}

\title{Adaptive Testing of Computer Vision Models}

\author{Irena Gao\\
Stanford University\thanks{Undertaken in part as an intern at Microsoft Research.}\\
{\tt\footnotesize irena@cs.stanford.edu}
\and
Gabriel Ilharco\\
University of Washington\\
{\tt\footnotesize gamaga@cs.washington.edu}
\and
Scott Lundberg and Marco Tulio Ribeiro\\
Microsoft Research\\
{\tt\footnotesize \{marcotcr, scott.lundberg\}@microsoft.com}
}
\maketitle

\begin{abstract}
Vision models often fail systematically on groups of data that share common semantic characteristics (e.g., rare objects or unusual scenes), but identifying these failure modes is a challenge. 
We introduce \tool, an interactive process for testing vision models which helps users identify and fix coherent failure modes.
Given a natural language description of a \slice, \tool{} retrieves relevant images from \laion{} with \clip.
The user then labels a small amount of data for model correctness, which is used in successive retrieval rounds to hill-climb towards high-error regions, refining the group definition. 
Once a group is saturated, \tool~uses \gpt{} to suggest new group descriptions for the user to explore.
We demonstrate the usefulness and generality of \tool{} in user studies, where users find major bugs in state-of-the-art \class, \detc, and \capt{} models.
These user-discovered groups have failure rates 2-3x higher than those surfaced by automatic error clustering methods.
Finally, finetuning on examples found with \tool{} fixes the discovered bugs when evaluated on unseen examples, without degrading in-distribution accuracy, and while also improving performance on out-of-distribution datasets.
\vspace{-0.4cm}

\end{abstract}


\begin{figure*}[t]
    \centering
    \includegraphics[width=\textwidth]{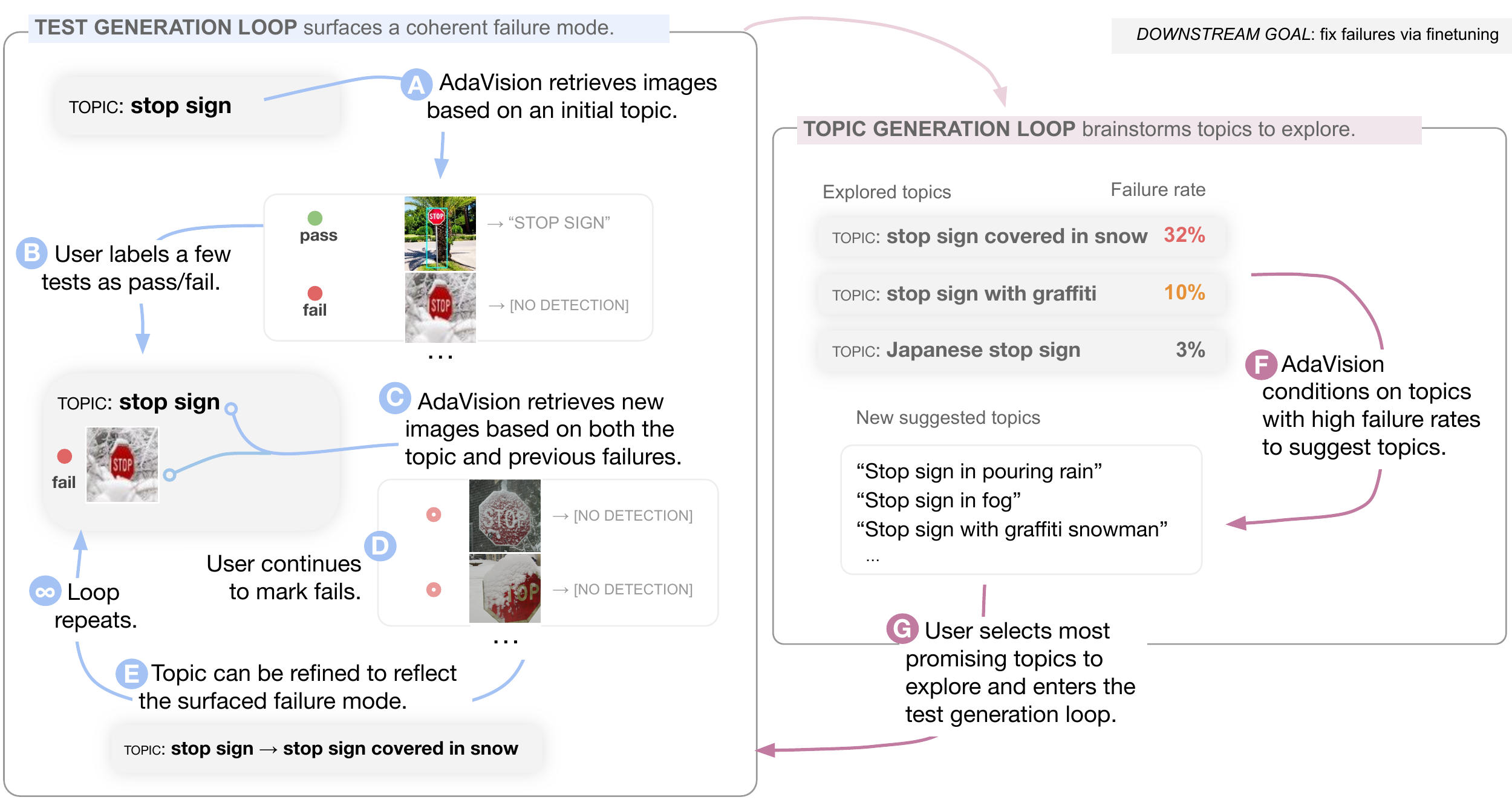}
    \caption{\tool~is a human-in-the-loop tool for surfacing coherent groups of failures, which are indexed via natural language \textit{topics}. In the \inl~(left), \tool~generates challenging tests for a topic, hill-climbing on previous failures. In the \outl~(right), \tool~generates new topics to explore, hill-climbing on previously difficult topics. Users steer testing by labeling a small number of images in the \inl~and selecting which topics to explore from the \outl.\vspace{0cm}}
    \label{fig:pull}
\end{figure*}

\section{Introduction}\label{sec:intro}
Even when vision models attain high average performance, they still fail unexpectedly on \subpops~of images.
When low-performing \subpops~are \textit{semantically coherent} (\ie unified by a human-understandable concept), their identification helps developers understand how to intervene on the model (\eg by targeted data collection) and decide if models are safe and fair to deploy~\cite{eyuboglu2022domino,mitchell2019model}.
For example, segmentation models for autonomous driving fail in unusual weather. 
Because we have identified this, we know to deploy such systems with caution and design interventions that simulate diverse weather conditions~\cite{tian2018deeptest,zendel2018wilddash}.
Identifying coherent failure modes helps developers make such deployment decisions and design interventions.

However, discovering coherent error groups is difficult in practice, since most evaluation sets lack the necessary visual or semantic annotations to group errors.
Prior work clusters evaluation set errors in different representation spaces~\cite{d2022spotlight,eyuboglu2022domino,jain2022distilling,sohoni2020no,wiles2022discovering}, but these methods often produce incoherent groups, such that it is hard for humans to assess their impact or fix them.
These methods are also limited by the coverage of small evaluation sets, which underestimate out-of-distribution vulnerabilities \cite{patel2008investigating, imagenettoimagenet,ribeiro2020beyond}, and become less useful as models approach near-perfect accuracy on benchmarks.
An alternative approach for discovering failures is open-ended \textit{human-in-the-loop testing}~\cite{ganguli2022red,ribeiro2022adaptive,ribeiro2020beyond}, which leverages interaction with users to generate challenging data to test models on coherent topics.
While successful in NLP, there are no established frameworks for open-ended testing in vision.
In this work, we present \textbf{Ada}ptive Testing for \textbf{Vision} Models (\tool), a process and tool for human-in-the-loop testing of computer vision models.
As illustrated in Figure \ref{fig:pull} (left), a user first proposes a \slice~of images to evaluate using natural language (e.g. \topic{stop sign}). 
This description is used to retrieve images from a large unlabeled dataset (\laion) using \clip{} embeddings \cite{radford2021learning}.
After users label a small number of the returned images for model correctness (pass / fail), the tool adapts to retrieve images similar to the discovered failures (Figure \ref{fig:pull}C).
\tool~reduces the manual labor required for human-in-the-loop testing by automatically hill-climbing towards high-error regions, while having a human-in-the-loop ensures groups are coherent and meaningful for the downstream application.
\tool~also leverages a large language model (\gpt~\cite{gpt3}) to adaptively help users generate descriptions for challenging groups to explore, as previously proposed by~\cite{ribeiro2022adaptive} and illustrated in Figure \ref{fig:pull} (right).
After testing, users finetune their models on discovered groups to \emph{fix} the bugs, and they can test again to verify improvement.

We demonstrate the usefulness of \tool{} in user studies, where users found a variety of bugs in state-of-the-art \class, \detc, and \capt{} models.
\tool{} groups had failure rates 2-3x higher than those found by \domino{}, an automatic error clustering baseline~\cite{eyuboglu2022domino}.
Further, users found close to 2x as many failures with \tool{}, when compared to a strong non-adaptive baseline using the same \clip{} backend.
Finally, we show that finetuning a large classification model on failures found with \tool~improves performance on held-out examples of such groups \emph{and} on out-of-distribution datasets, without degrading in-distribution performance: 
finetuning an ImageNet-pretrained \vit~model \cite{deng2009imagenet,dosovitskiy2020image} on user study data fixes the discovered groups (boosting accuracy from 72.6\% to 91.2\%) without reducing overall accuracy on ImageNet, while also improving the accuracy of labeled classes (78.0\% to 84.0\%) on five out-of-distribution (OOD) ImageNet evaluation sets.

\section{Related Work}\label{sec:related_work}
\tightparagraph{Automatic group discovery.}
To help humans find low-performing \slices, one line of prior work clusters errors in validation data, labeling each cluster with a caption~\cite{d2022spotlight,eyuboglu2022domino,jain2022distilling,sohoni2020no,wiles2022discovering}.
A desirable property for these clusters is \emph{coherency}: groups and captions that are semantically meaningful to humans aid decisions about safe deployment and intervention (\eg collecting more data to fix the bugs).
Further, clusters should \emph{generalize}: since each cluster is meant to represent a bug in the model, collecting more data matching the caption should result in a high failure rate.
Prior work finds that automatic methods which cluster validation set errors can fail this second criterion: clusters can spuriously overfit to a few mispredicted examples \cite{johnson2023does}.
Overfitting is particularly likely on small or mostly-saturated evaluation sets.
In contrast, \tool{} leverages a human in the loop to iteratively test models, encouraging descriptions which are coherent, generalizable, and relevant for the users' task.

\tightparagraph{Testing machine learning models.}
Human-aided \emph{testing} of models is an established practice in Natural Language Processing~\cite{bhatt2021case,ganguli2022red,kiela2021dynabench,ribeiro2022adaptive,ribeiro2020beyond}.
This area applies insights from software engineering by having users \emph{create} test cases with templates \cite{ribeiro2020beyond}, via crowdsourcing \cite{ganguli2022red, kiela2021dynabench}, or with help from a language model \cite{ribeiro2022adaptive}.
Tests are organized into \slices~and used to evaluate a target model.
This style of testing, which leverages human steering to probe inputs \textit{beyond} traditional training / validation splits, has successfully unearthed coherent bugs in state-of-the-art NLP models, even as models saturate static benchmarks~\cite{ganguli2022red,kiela2021dynabench,ribeiro2022adaptive,ribeiro2020beyond}.

In contrast, testing in computer vision has not moved far from static evaluation sets, with testing limited to pre-defined suites of data augmentations~\cite{du2022vision,tian2018deeptest,zhang2018deeproad}, static out-of-distribution test sets~\cite{objectnet,imagenetr,imageneta,recht2019imagenet,imagenetsketch}, training specific counterfactual image generators~\cite{balakrishnan2021towards,denton2019image,khorram2022cycle}, or using 3D simulation engines~\cite{bogdoll2022one,leclerc20213db}.
All of these methods either restrict tests to a static set of images, or along pre-specified axes of change (\eg blur augmentations), and many introduce synthetic artifacts.
In contrast, \tool~enables dynamically testing models along unrestricted axes by allowing users to specify tests using natural language.
Moreover, \tool~can pull images from 5 billion total candidates, orders of magnitude larger than typical evaluation datasets.

Our work shares motivations with prior work that compares models via dynamically selected test sets \cite{ma2018group,wang2020going,wang2008maximum,yan2021exposing} and with concurrent work by Wiles et al.~\cite{wiles2022discovering}, who also leverage foundation models for open-ended model testing of computer vision models.
Like other automatic methods, their approach involves clustering evaluation set errors, captioning these clusters, and then generating additional tests per cluster using a text-to-image generative model~\cite{saharia2022photorealistic}. 
\tool~differs in that it is human-in-the-loop; as in prior work, we find that a small amount of human supervision, which steers the testing process towards meaningful failures for the downstream application, is effective at identifying coherent bugs~\cite{ribeiro2022adaptive} and avoids the pitfalls of automatic group discovery from evaluation sets (our discovered bugs have failure rates orders of magnitude higher than Wiles et al.~\cite{wiles2022discovering}).

\begin{figure*}[bt]
    \centering
    \includegraphics[width=0.915\textwidth]{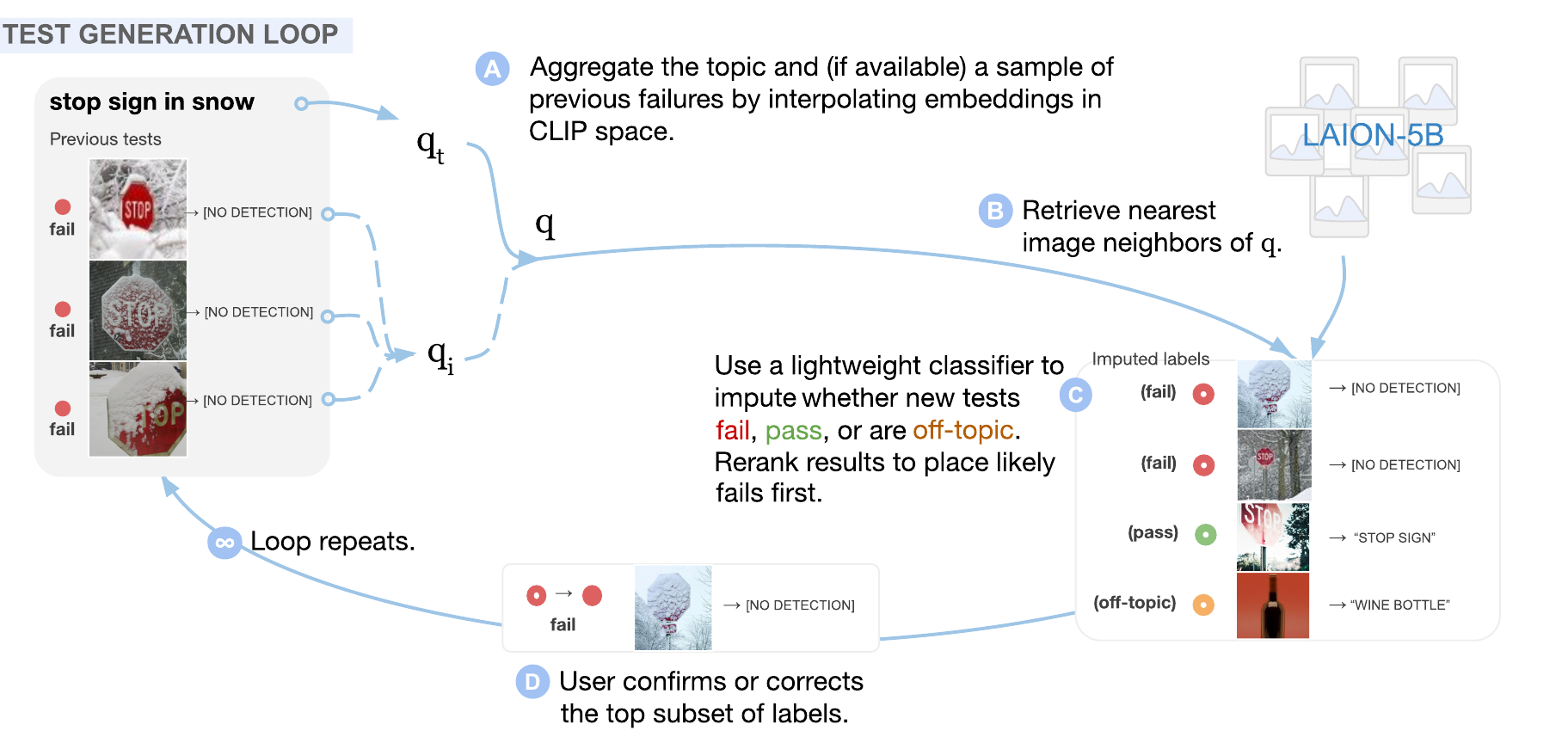}
    \caption{
        In the \inl, \tool~populates a topic with image tests, hill-climbing on previous failures through embedding interpolations. To minimize user labeling effort, \tool~also uses lightweight classifiers to automatically sort and label returned tests. We provide additional technical details on these steps in Appendix A.1.
        \vspace{0.2cm}
    }
    \label{fig:inl}
\end{figure*}

\section{Methodology}\label{sec:method}

We aim to test vision models across a broad set of tasks, including \class, \detc, \capt.
Given a model $\model$, we define a \textbf{test} as an image $x$ and the expected behavior of $\model$ on $x$~\cite{ribeiro2022adaptive,ribeiro2020beyond}. 
For example, in object recognition, we expect that $m(x)$ outputs one of the objects present in $x$, while in captioning, we expect $m(x)$ to output a factually correct description for $x$. 
A test fails if $m(x)$ doesn't match these expectations.

A \slice{}, or \textbf{topic} \cite{ribeiro2022adaptive}, contains tests whose images are united by a human-understandable concept \cite{eyuboglu2022domino,wiles2022discovering}) and by a shared expectation \cite{ribeiro2020beyond}.
\tool's goal is to help users discover topics with high failure rates, henceforth called \textbf{bugs} \cite{ribeiro2022adaptive,wiles2022discovering}.
Assuming a distribution of images given topics $P(X|T)$, a bug is $t \in T$ such that failure rates are greatly enriched over the baseline failure rate:
$$
\mathbb{E}_{x \sim \mathbf{P(X|t)}} \left[ \text{test}(x)  \text{ fails} \right] \gg \mathbb{E}_{x \sim \mathbf{P(X)}} \left[ \text{test}(x) \text{ fails} \right]
$$

For a given topic, users start with a textual topic description (e.g. \topic{stop sign} in Figure \ref{fig:pull} left), and then engage in the \textit{\inl} (Section \ref{sec:inner_loop}), where \tool{} generates test suggestions relevant to the topic. At each iteration, \tool~adaptively refines the topic based on user feedback on topic images, steering towards model failures.
While users can explore whatever topics they choose (e.g. based on the task labels, application scenarios, or existing topics from prior testing sessions), \tool{} also includes a \textit{\outl} (Figure \ref{fig:pull} right; Section \ref{sec:outer_loop}) where a large language model suggests topics that might have high failure rate, based on existing topics and templates.
At the end of the process, users accumulate a collection of topics and can then intervene on identified bugs, \eg by finetuning on the failed tests to improve performance on fresh tests $x \sim P(X|t)$ from the topic (Section \ref{sec:finetuning}).



\subsection{\Inl}\label{sec:inner_loop}
In the \inl, users explore a candidate topic $t$.
At each iteration, users get test suggestions and provide feedback by labeling tests, changing the topic name, or both. This feedback adaptively refines the definition of $t$, such that the next round of suggestions is more likely to contain failures (Figure \ref{fig:pull} left).


\tightparagraph{Initial test retrieval.} 
Given a topic string $q$, \tool~retrieves a warm-up round of tests (Figure \ref{fig:pull}A) by using the text embedding $q_t$ (embedded with \clipvit~\cite{radford2021learning}) to fetch nearest image neighbors from \laion{}~\cite{schuhmann2022laion}, a 5-billion image-text dataset.\footnote{We use \url{https://github.com/rom1504/clip-retrieval}}
We note that \laion{} can be replaced by or supplemented with any large unlabeled dataset, or even with an image generator.


\tightparagraph{Adaptive test suggestions.}
We run the target model on the warm-up images, obtaining $(x, \model(x))$ tuples.
Users then label a small number of these tests as \textit{passed}, \textit{failed}, or \textit{off-topic}.
A test is off-topic if it is a retrieval error  (\eg not a stop sign in Figure \ref{fig:inl}C), or if the test is not realistic for the downstream application.
When labeling, users prioritize labeling failures.
We incorporate these labeled tests in subsequent rounds of retrieval, where we suggest tests based both on the textual description (\topic{stop sign}) and visual similarity to previous failures.
To do so (Figure \ref{fig:inl}A, B), we sample up to 3 in-topic images (prioritizing failures), combine their embeddings into a single embedding $q_i$ using a random convex combination of weights, and generate a new retrieval query by spherically interpolating each $q_i$ with the topic name embedding $q_t$, as done in~\cite{ramesh2022hierarchical}.\footnote{q = $\mathrm{slerp}(q_i, q_t) = \frac{\sin((1-\lambda)\alpha)}{\sin\alpha}q_i + \frac{\sin(\lambda\alpha)}{\sin\alpha}q_t$, where \\$\cos\alpha=\langle q_i, q_y \rangle$. We sample $\lambda \sim \text{Uni}(0,1).$} 
We automatically filter retrievals to prevent duplicate tests.
We provide more technical details in Appendix A.1.

By incorporating images into retrieval, \tool~adaptively helps users refine the topic to a \slice~of failures. 
Each round can be seen as hill-climbing towards a coherent, high-error region, based on user labels.
We evaluate the effectiveness of this strategy in Section \ref{sec:ablation}, where we observe that it significantly improves retrieval from \laion{}.

\tightparagraph{Automatically labeling tests.}
In order to minimize user labeling effort, we train lightweight topic-specific classifiers to re-rank retrieved results according to predicted pass, fail, or off-topic labels (Figure \ref{fig:inl}C).
For each topic, we take user pass/fail labels and train a Support Vector Classifier (SVC) on concatenated \clip{} embeddings of each test's input (image) and output (\eg predicted label).
If off-topic labels are provided, we train a second SVC model to predict whether a test is in-topic or off-topic.
The predictions of these two models are used to rerank the retrievals such that likely failures are shown first (sorted by the distance to the decision boundary), and tests predicted as off-topic are shown last. The user also sees a binary prediction of pass / fail (Figure \ref{fig:inl}C), so they can skip tests predicted as ``pass'' once the lightweight models seem accurate enough.
These models take less than a second to train and run, and thus we retrain them after every round of user feedback.

\subsection{\Outl}\label{sec:outer_loop}
In the \outl~(Figure \ref{fig:pull} right), users collaborate with \tool~to generate candidate topics to explore.
While labeling examples in the \inl{} is easy for humans, generating new topics is challenging, even when users are tasked with testing $\model$ for a single concrete label (e.g. \topic{stop sign}).
Thus, we offload this creative task to a large language model (\gpt, text-davinci-002), inspired by successes in related NLP tasks \cite{ribeiro2022adaptive}.

\begin{figure}[tb]
    \centering
    \includegraphics[width=0.455\textwidth]{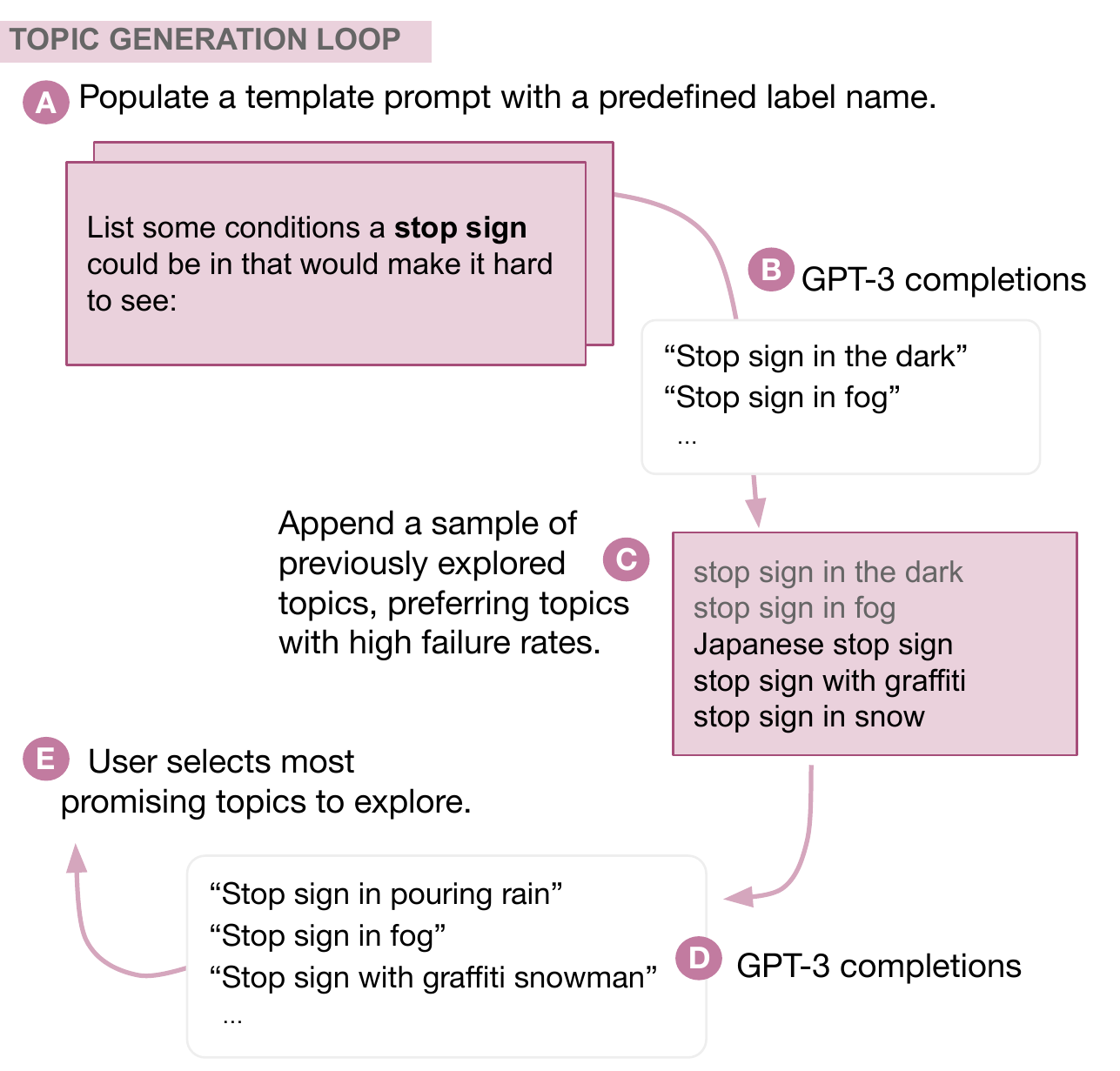}
    \caption{
        In the \outl, \tool~leverages \gpt~to generate topics for users to explore. These suggestions condition on previously explored topics with high failure rates.     
        \label{fig:outl}
    }
\end{figure}

As illustrated in Figure \ref{fig:outl}, we start by using a collection of prompt templates,
such as  {``List some conditions a \{LABEL\} could be in that would make it hard to see''} and {``List some unusual varieties of \{LABEL\}''}, replacing \{LABEL\} at testing time with predefined label names or existing user topics (e.g. \topic{stop sign}).
We combine completions of this prompt with existing user topics (prioritizing topics with high failure rates) into a new few-shot prompt, such that GPT-3 is ``primed'' to return high-failure topic names \cite{ribeiro2022adaptive}.
The resulting topic name suggestions are presented to the user, who chooses to explore topics they deem interesting and important.
These suggestions only need high recall (not precision), as users can disregard irrelevant suggestions.

\section{Evaluation}\label{sec:evaluation}
To evaluate \tool, we first quantify the value of \textit{adaptive} test suggestions   (\ie retrieving tests using interpolated topics and images, Section \ref{sec:inner_loop}) for finding failures. 
Then, we verify that \tool~helps users find coherent bugs in state-of-the-art vision models across a diverse set of tasks in a set of user studies (Section \ref{sec:user_studies}).
These also demonstrate that \tool~is more effective than a non-adaptive version relying on an interactive CLIP search. 
In a separate experiment, we compare \tool~and \domino, an automatic slice discovery method (Section \ref{sec:domino}). 
Finally, we use finetuning to patch the discovered bugs (Section \ref{sec:finetuning}), improving performance in these topics.

\subsection{Value of adaptive test suggestions}\label{sec:ablation}
We ran a controlled experiment to understand the value of the \inl's adaptivity for finding failures.
We compared the number of failures found within a topic when using adaptive test suggestions, compared to retrieving based on topic name alone.
To do so, we fixed a set of broad topics from two tasks (\class~and \detc) and labeled the top 100 retrievals found by each strategy.\footnote{One of the authors labeled all images in this experiment.}
For \class, we created six broad topics with the template \topic{a photo of a \{y\}} with the labels \sixoverlap.\footnote{We selected these classes because they overlap on various ImageNet OOD datasets (ImageNet V2, ImageNet-A, \etc), discussed in Section \ref{sec:finetuning}.} 
For \detc, we use the template \topic{a photo of a \{y\} on the road} with labels \{cyclist, motorcycle, car, stop sign, person, animal\}.

\begin{figure}[tb]
    \centering
    \includegraphics[width=0.485\textwidth]{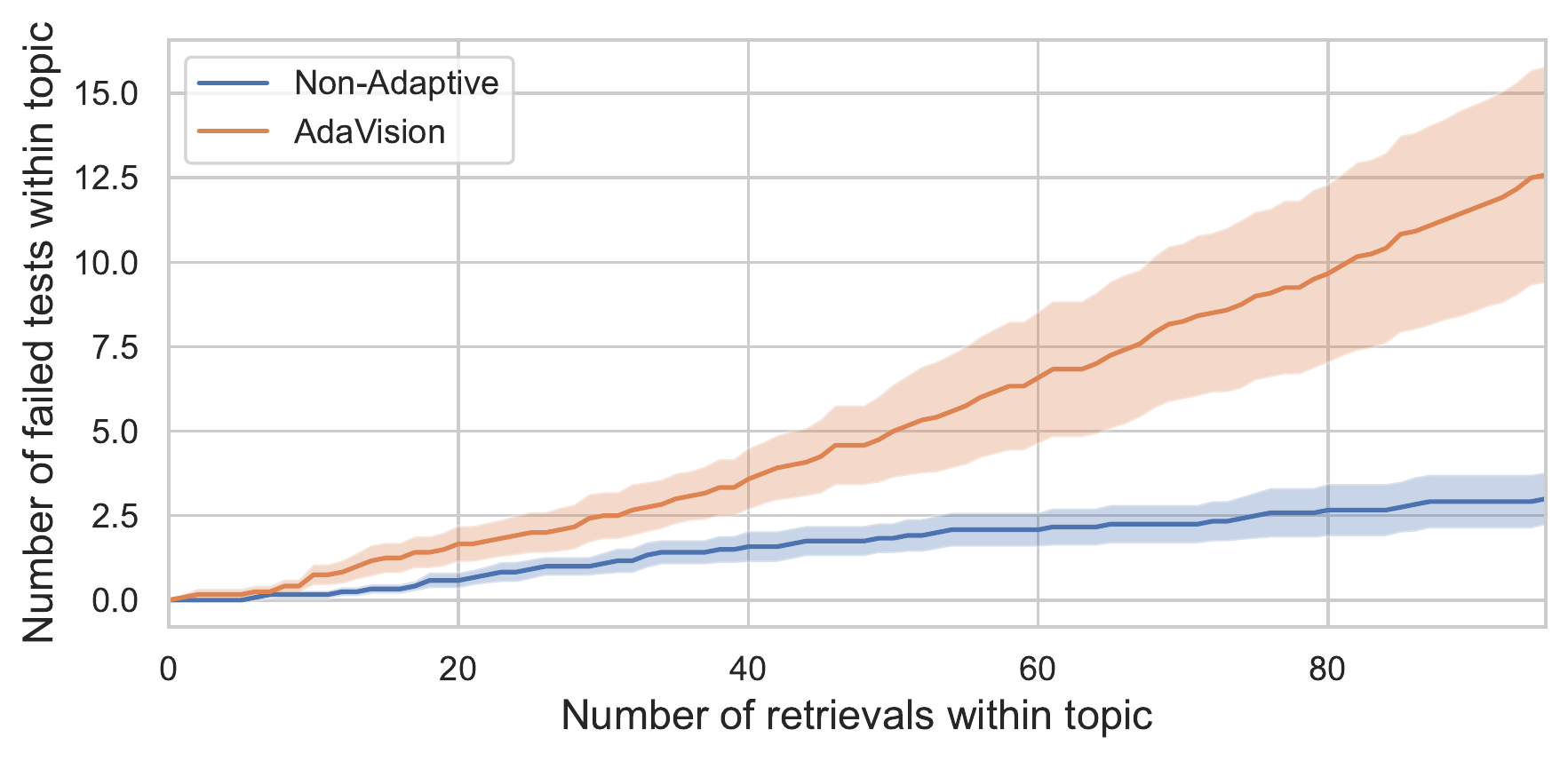}
     \caption{Average number of failures accumulated within a topic over the course of 100 retrievals, comparing \tool~with a non-adaptive baseline. \tool~is significantly more effective at finding failed tests, because it is able to quickly surface more failures once a few are found. Standard error is over 12 topics.}
     \label{fig:ablation}
\end{figure}

Figure \ref{fig:ablation} shows the number of failures found by \tool~compared to \baseline~over time, averaged across topics.
Once a small number of failures have been found, \tool~is able to quickly surface more failures, outperforming retrieval that only uses the topic string.
Even though these broad initial topics result in low baseline failure rates, \tool~surfaces \slices~of failures within the broad topic by hill-climbing on previous failures.

\subsection{User study}\label{sec:user_studies}
We ran user studies to evaluate whether \tool~enables users to find bugs in top vision models. 
Users are able to find coherent bugs with \tool~in state-of-the-art models, even though these models have very high in-distribution accuracy.
We also show that \tool's \textit{adaptivity}, \ie its hill-climbing on previous failures (both test and topic), helps users find nearly 2x as many failures than without adaptivity.

\tightparagraph{Tasks and models.}
To highlight the flexibility of \tool, we had users test models across three vision tasks (\class, \detc, and \capt).
We targeted models and categories with high benchmark or commercial performance, where failures are not easy to find, and we instructed users to use stringent definitions for model failure.
For \class, users tested \vit{} on two ImageNet categories \textit{banana} and \textit{broom} (chosen for their high top-1 accuracy of 90\%), and were instructed that a prediction that includes \textit{any} object in the image is counted as a valid prediction.
For \detc, users tested Google Cloud Vision API's Object Detection on two categories relevant for autonomous driving: \textit{bicycle} and \textit{stop sign} (average precision 0.7-0.8 on OpenImages).\footnote{\scriptsize \url{https://modelcards.withgoogle.com/object-detection}} Users were instructed to only mark as failures tests where the model does not detect \emph{any} bicycles or stop signs present.
For \capt, users tested Alibaba's official checkpoint of \ofa~finetuned on COCO Captions~\cite{wang2022unifying}, which is state-of-the-art on the benchmark, and were asked to explore scenes a visually impaired user might encounter when inside a \textit{kitchen} or an \textit{elementary school}. Users were instructed to consider as failures only object and action recognition errors which would egregiously mislead a visually impaired user.

\tightparagraph{Participants and setup.}
We recruited $40$ participants from academia and industry (with IRB approval) who had taken at least a graduate-level course in machine learning, computer vision, or natural language processing. We assigned 16 users to the classification task, 16 to the detection task, and 8 users to the image captioning task. 

In these studies, we also aimed to ablate the importance of \tool's {adaptivity} over its benefits as an interactive search interface with model scoring. 
To do so, we asked each user to complete two rounds of testing. 
In the \tool~round, users had full access to \tool~as described in Section \ref{sec:method}, while in \baseline{} round we disabled topic suggestions, automatic test labeling, and adaptive test suggestions (i.e. suggestions are always retrievals based solely on the topic name).
Users had a limited amount of time for each round (15 to 20 minutes), and were instructed to try to find as many failure-prone topics (bugs) for a specific category as possible, switching topics whenever they found 8-10 failures within a topic (more details in Appendix B).
Users tested different categories between rounds (to minimize learning between rounds), and category assignments and round orderings were randomized.

\begin{figure}[!btp]
    \centering
    \includegraphics[width=0.475\textwidth]{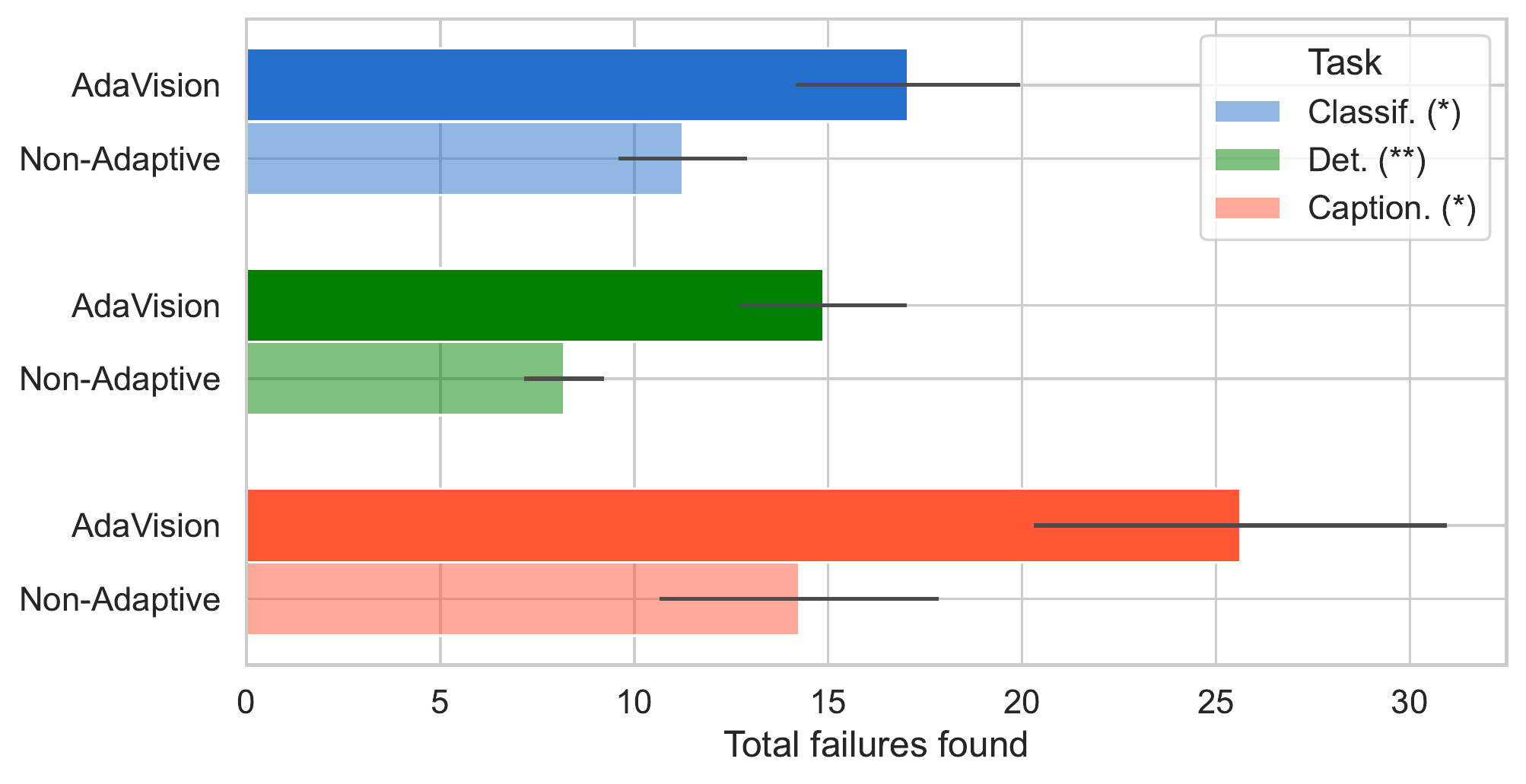}
    \caption{
        User study results comparing \tool~to \baseline~(baseline). 
        Error bars are standard errors over users. 
        Results significant with $p < 0.05$ (*) or $p < 0.005$ (**), with more details in Appendix B.
    }
    \label{fig:user_studies}
\end{figure}

\begin{figure*}[!tb]
    \centering
    \includegraphics[width=\textwidth]{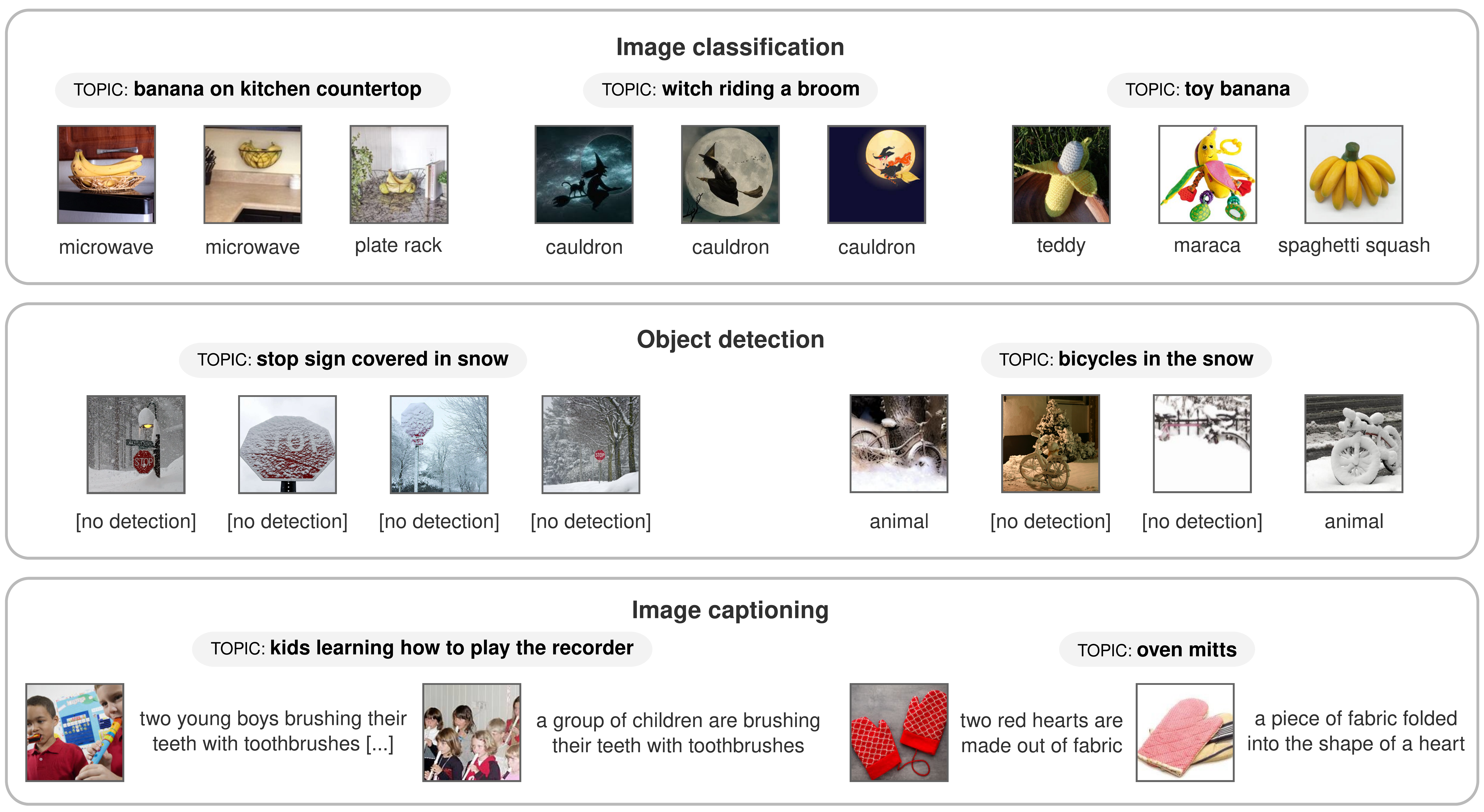}
    \caption{Sample of bugs found by users in studies. In each case, the model prediction (shown to the right of the corresponding input image) is incorrect. These bugs span spurious correlations (e.g. \vit{} associating a kitchen counter top with a microwave), difficult examples (e.g. Cloud Vision API failing to detect stop signs partially obscured by snow), and missing world knowledge (e.g. \ofa~misidentifying oven mitts).}
    \label{fig:samples}
    \vspace{0.1cm}
\end{figure*}

\tightparagraph{Results.}
We present the number of failures found (averaged across users) in Figure \ref{fig:user_studies}. 
Users were able to find a variety of bugs with \tool, even in strong models with strict definitions of failure.
Further, \tool's adaptivity helped users find close to twice as many failing tests than \baseline{}, with moderate to large (standardized) effect sizes in \class~($d=0.588, p < 0.05$), \detc~($d=0.882, p < 0.005$), and \capt~($d=0.967, p < 0.05$).
Using \tool helped users identify \emph{more diverse} bugs than the baseline: while 12/40 users found 2 or more bugs with \tool, only 1/40 could match this level of diversity in the baseline round.

Qualitatively, users found bugs related to spurious correlations, difficult examples, and missing world knowledge (we share samples in Figure \ref{fig:samples}). For example, users discovered that \vit{} strongly correlates kitchen countertops with the label \textit{microwave} and witch hats with the label \textit{cauldron}, leading to failure on images where these correlations do not hold (\eg microwaves are absent). 
Users found that Cloud Vision misses detections when stop signs and bicycles are partially obscured by snow, and users also discovered object and action recognition errors in \ofa, such as with \textit{oven mitts} and musical instruments held near the mouth.

We surveyed users on whether \tool~was instrumental in finding these bugs. 84.6\% of users marked said they ``could not have found these bugs using existing error analysis tools [they] have access to.''
We also asked users to rate the cognitive difficulty of finding bugs in each round, on a scale of 1 to 5.
The average perceived difficulty with \tool~was $3.05 \pm 1.07$, in contrast with $4.10 \pm 0.91$ for \baseline{}. In a paired t-test, this gap was significant with $p< 0.001$, reflecting that users felt testing was easier with \tool~than without.


\subsection{Comparison with automatic slice discovery}\label{sec:domino}
\domino{} \cite{eyuboglu2022domino} is a state-of-the-art slice discovery method that clusters validation set errors and describes them with automatically generated captions.
We compare these to \tool{} topics on unseen data, noting that if a topic or caption $t$ genuinely describes a bug, drawing new samples from $x \sim P(X|t)$ should yield a high failure rate.

\tightparagraph{Setup.}
We compare bugs found in ImageNet classification models with respect to six categories: \sixoverlap.
Specifically, tests failed if they were false negatives: for class $y$, the model fails if the image $x$ contains object $y$, but the model instead predicts an object that is not in the image.
We target the \vit~model from Section \ref{sec:user_studies} \cite{dosovitskiy2020image}, and a \resnet \cite{he2016deep}. 

For each category, \domino~clusters ImageNet validation examples using an error-aware Gaussian mixture model in \clip's latent space, and then describes each cluster with a caption.\footnote{We use the official implementation \cite{eyuboglu2022domino}, available at\\ \url{https://github.com/HazyResearch/domino}}
We use two variants, \dominobert~and \dominoofa, which differ in how they caption clusters (template filling with BERT \cite{devlin-etal-2019-bert} or captioning with OFA \cite{wang2022ofa}). Appendix C contains more details.

We used \tool{} topics from a user session in the user studies for overlapping categories, and had an author run additional sessions (i.e. use \tool{} for 20 minutes) for the 2 remaining categories.
While \domino{} targets each model individually, we only target \vit{}, and directly transfer the discovered topics to \resnet.

Both methods propose five topics per category (we took the top-5 with most failures for \tool{}), for a total of 30 topics each (listed in Appendix C).
To evaluate the failure rate of a topic on \emph{new} data that matches the topic description, we retrieve nearest neighbors from \laion{} using the topic name, and label the first 50 in-topic retrievals (skipping over images that do not fit the topic description).
We exclude tests \tool{} users encountered during testing to avoid counting tests already reviewed in \tool's favor. 

\tightparagraph{Results.}
We present average failure rates across coherent topics proposed by each method in Table \ref{tab:automatic_eval}.
We also present two baselines: the failure rate of a generic topic description per category (\topic{a photo of \{y\}}), and the failure rate on the original ImageNet validation data (noting that ImageNet has a looser definition of failure, enforcing prediction on an arbitrary object on images with multiple objects).
\tool{} topics yield much higher failure rates, while \domino{} shows rates close to baselines. Interestingly, \tool{} topics generated while testing \vit{} transfer well to \resnet{}.

Further, while all \tool{} topics are unsurprisingly coherent (as they are human-verified), we find that 61.6\% and 33.3\% of topics from \dominobert{} and \dominoofa{} respectively are nonsensical (\eg \topic{a photo of setup by banana}, \topic{a photo of skiing at sandal}) or fail to refer to the target category at all (\eg \topic{three oranges and an apple on a white background} when the target is ``lemon''). These topics are excluded from Table \ref{tab:automatic_eval} and listed in Appendix C.

We believe these results illustrate some inherent difficulties of automatic slice discovery methods. 
Validation error clusters may not be semantically tied together, especially when models saturate in-distribution benchmarks (\vit{} and \resnet{} are stronger than models used in prior evaluation of automatic slice discovery~\cite{eyuboglu2022domino,jain2022distilling}).
Current slice captioning methods may also over-index into incorrect details or miss broader patterns between images.
Because of the human-in-the-loop, \tool~enables users to form more coherent hypotheses about model failures.
Further, cluster captions can describe a group of failures without including the \textit{cause} of the model failure \cite{johnson2023does} (e.g. ``a woman sitting on a chair holding a broom'' is coherent, but \vit{} has a failure rate of only 10\% on additional data matching this description). 
In contrast, users of \tool{} iteratively form hypotheses about model vulnerabilities: 
after selecting a topic, users observe model behavior on additional data from the topic, 
leading them to \emph{refine} the topic definition.
This iterative process helps users identify topics which consistently capture model failures.

\begin{table}[bt]
    \centering
    \footnotesize
    \begin{tabular}{llc}
    \toprule
    \textbf{Model}            & \textbf{Method} & \textbf{Avg failure rate} \\ \midrule
    \multirow{7}{*}{ViT-H/14} & \textit{\topic{a photo of \{y\}}}    & 1.33                                                             \\ \cmidrule(l){2-3} 
                              & \textit{ImageNet}             & 11.47                                               \\ \cmidrule(l){2-3} 
                              & \dominobert                   & 8.6                                              \\ \cmidrule(l){2-3} 
                              & \dominoofa                    & 7.33                                                    \\ \cmidrule(l){2-3} 
                              & \textbf{\tool}                & \textbf{28.47}                                         \\ \midrule
    \multirow{7}{*}{ResNet50} & \textit{\topic{a photo of \{y\}}}    & 15.7                                                   \\ \cmidrule(l){2-3} 
                              & \textit{ImageNet}             & 23.67                                                         \\ \cmidrule(l){2-3} 
                              & \dominobert                   & 20.44                                                        \\ \cmidrule(l){2-3} 
                              & \dominoofa                    & 25.45                                                     \\ \cmidrule(l){2-3} 
                              & \textbf{\tool}                & \textbf{56.93}                                                \\ \bottomrule
    \end{tabular}%
    \caption{Average failure rates across topics proposed by \tool~and two variants of \domino, an automatic slice discovery method. \tool~finds bugs that are ~3x more difficult for models, while automatic methods propose groups that are close to baseline failure rates.
    \label{tab:automatic_eval}}
\end{table}



\begin{table*}[h]
    \centering
    \small
    \setlength\tabcolsep{8.5pt}
    \resizebox{\textwidth}{!}{%
    \begin{tabular}{@{}llllllll@{}}
    \toprule
    \textbf{}                 & \multicolumn{3}{c}{{\tool~Topics}}       & \multicolumn{1}{c}{{ImageNet}}                 & \multicolumn{2}{c}{{Avg across  OOD Eval Sets}} \\ \midrule
    \textbf{Model}            & \multicolumn{2}{l}{\textbf{Treatment Topics}}  & \textbf{Control Topics} &  \textbf{Overall} & \textbf{Treatment Classes}  & \textbf{Overall} \\ 
                              & \multicolumn{1}{l}{\textit{\footnotesize \laion}}             &  \multicolumn{1}{l}{\textit{\footnotesize Google}}                  &                                     &                    &                            & \\
    \midrule
    Before finetuning         & 72.6      & 76.7               & 91.3                   &  88.4                    & 78.0                        & 77.7                     \\
    Finetuning with \textit{\topic{an image of \{y\}}} & 82.5 (0.9)        & 82.9 (0.6)              & 90.8 (0.3)                   &   \textbf{88.5 (0.0)}                        & 82.1 (0.6)                  &     78.0 (0.1)                     \\
    Finetuning with \tool tests & \textbf{91.2 (0.5)}    & \textbf{90.6 (0.6)}             & \textbf{91.9 (0.2)}                    &             88.4 (0.0)              & \textbf{84.0 (0.2)}                  &    \textbf{78.2 (0.0)}                     \\ \bottomrule
    \end{tabular}%
    }
    
    \caption{Accuracies on \tool~topics, ImageNet \cite{deng2009imagenet}, and five OOD ImageNet evaluation sets \cite{objectnet,imagenetr,imageneta,recht2019imagenet,imagenetsketch} before and after finetuning on images accumulated from testing with~\tool. Compared to a baseline of finetuning on the same number of images pulled generically using the topic \topic{an image of \{y\}} from \laion, \tool~improves accuracy on held-out data from topics in the finetuning set (left two columns), regardless of whether images are sourced from \laion~or Google Images. \tool~also improves accuracy on OOD evaluation sets (right two columns). Finetuning maintains overall performance on ImageNet (center) and held-out control topics (third column).
    \label{tab:finetuning}}
    \vspace{0.1cm}
\end{table*}


\subsection{Fixing bugs via finetuning}\label{sec:finetuning}

We evaluate whether users can fix bugs discovered with \tool, by finetuning on failed tests.
We finetune \vit{} on the $30$ \tool{} topics for categories \sixoverlap{} from Section \ref{sec:domino}, taking a sample of 20 tests per topic for a total of $600$ images.
As a baseline that just trains on images from a different distribution, we finetune \vit{} on an equal-size set of images retrieved from \laion~using generic topics in the form \topic{an image of \{y\}}.
We evaluate the finetuned models on held-out examples from \tool~topics, on the original domain (ImageNet \cite{deng2009imagenet}), and on five out-of-distribution datasets: ImageNet V2~\cite{recht2019imagenet}, ImageNet-A~\cite{imageneta}, ImageNet-Sketch~\cite{imagenetsketch}, ImageNet-R~\cite{imagenetr}, and ObjectNet~\cite{objectnet}.
Additional details are in Appendix D.

\tightparagraph{Finetuning improves performance on discovered bugs.} We use $50$ held-out examples drawn from each topic we attempted to fix (\textit{treatment topics}) to verify if the bugs were patched. 
As shown in Table \ref{tab:finetuning}, finetuning substantially increases performance on the held-out data from the treatment topics (by 18.6 percentage points), making the accuracy on treatment topics surpass average accuracy on ImageNet. 
These performance gains also hold on in-topic images sourced from \textit{outside} of the \laion~distribution: for each treatment topic, we collected 50 images from a Google Image Search, deduplicated against both the finetuning set and the \laion~evaluation set.
Finetuning on \tool~images from a testing session using \laion also improves performance on in-topic images from Google by 13.9 percentage points.

\tightparagraph{Finetuning maintains in-distribution accuracy and improves OOD accuracy.} 
To ensure performance gains are not due to the introduction of new shortcuts, we check performance on the original in-distribution data (ImageNet). 
Finetuning on the treatment topics does not reduce overall ImageNet accuracy (Table \ref{tab:finetuning}).
Additionally, finetuning with \tool{} {improves} overall average performance on the treatment classes across out-of-distribution evaluation sets (from 78\% to 84\%).
To check for shortcuts at a more fine-grained level, we evaluate performance on 19 semantically contrasting \textit{control} topics with different labels.
For example, one treatment topic involved images of lemons next to tea, which the model often predicted as \textit{consomme}.
We added \topic{consomme} as a control topic to check that the model does improve performance on lemons at the expense of the ``consomme'' concept.
Similarly, for topics \topic{banana on kitchen countertop} and \topic{witch riding a broom} in Figure \ref{fig:samples}, we add the control topics \topic{microwave in kitchen} and \topic{witch with cauldron}; see Appendix D for a list.
Performance on the control topics does not decrease after finetuning.
Our results indicate users can fix \emph{specific} bugs discovered with \tool{} in \vit~without degrading performance elsewhere.

\section{Limitations}\label{sec:limitations}
\tightparagraph{Retrieval limitations.}
While \laion{} has good coverage for everyday scenes, it may not be appropriate to specialized domains such as biomedical or satellite imagery, and \clip's representation power is also likely to deteriorate on these domains \cite{radford2021learning}.
Even for everyday scenes, the quality of \clip{}'s text-based retrieval degrades as the complexity of the topic name increases, particularly when several asymmetric relations are introduced~\cite{ma2022crepe,yuksekgonul2022and}.
Further work to improve image-text models like \clip~could reduce off-topic retrievals during testing, improving users' testing speed.

\tightparagraph{Experiment limitations.}
We show that finetuning a state-of-the-art \class{} model on \tool{} bugs fixes them without degrading performance elsewhere.
While it is particularly encouraging that we could improve performance on labels that had very high accuracy to begin with, this experiment was done with a limited set of labels and with only one round of testing.
Multiple rounds of testing / finetuning could be more beneficial~\cite{ribeiro2022adaptive}.
Models smaller than \vit~may also be more prone to catastrophic forgetting \cite{ramasesh2021effect}.
To preserve in-distribution performance when fixing bugs for these models, robust finetuning techniques like weight averaging \cite{wortsman2022robust} or adding in-distribution data to the finetuning set may be necessary.

\tightparagraph{Fixing non-classification bugs.} Classification tests contain the expected label implicitly in pass/fail annotations, and thus are easy to turn into finetuning data. However, the same is not true for detection or captioning tests, since we do not collect correct bounding boxes or captions during testing (only pass/fail annotations).
Fixing such bugs would require an additional step of labeling failing tests prior to finetuning, or using loss functions that explicitly allow for negative labels \cite{negative_labels}.

\section{Conclusion}\label{sec:conclusion}

We presented \tool, a human-in-the-loop process for testing vision models that mimics the life-cycle of traditional software development~\cite{ribeiro2022adaptive}.
By leveraging human feedback for models on vision tasks, \tool~helps users to identify and improve coherent vulnerabilities in models, beyond what is currently captured in-distribution evaluation sets.
Our experiments indicate the adaptive nature of \tool~improves the discovery of bugs,
and that finetuning on bugs discovered with \tool~boosts performance on the discovered failure modes.
\tool is open-sourced at \url{https://github.com/i-gao/adavision}.




\FloatBarrier

\section*{Acknowledgements}
We're grateful to 
Adarsh Jeewajee, Alexander Hoyle, Bhargavi Paranjape, Dhruba Ghosh, Mitchell Wortsman, Pang Wei Koh, Sarah Pratt, Shikhar Murty, Shiori Sagawa, and Tongshuang Wu
for giving feedback at various stages of this paper.

\vspace{2cm}
{\small
\bibliographystyle{ieee_fullname}
\bibliography{references}

\begin{thebibliography}{10}\itemsep=-1pt

\bibitem{balakrishnan2021towards}
Guha Balakrishnan, Yuanjun Xiong, Wei Xia, and Pietro Perona.
\newblock Towards causal benchmarking of biasin face analysis algorithms.
\newblock In {\em Deep Learning-Based Face Analytics}, pages 327--359.
  Springer, 2021.

\bibitem{objectnet}
Andrei Barbu, David Mayo, Julian Alverio, William Luo, Christopher Wang, Dan
  Gutfreund, Josh Tenenbaum, and Boris Katz.
\newblock Objectnet: A large-scale bias-controlled dataset for pushing the
  limits of object recognition models.
\newblock In H. Wallach, H. Larochelle, A. Beygelzimer, F. d\textquotesingle
  Alch\'{e}-Buc, E. Fox, and R. Garnett, editors, {\em Advances in Neural
  Information Processing Systems}, volume~32. Curran Associates, Inc., 2019.

\bibitem{bhatt2021case}
Shaily Bhatt, Rahul Jain, Sandipan Dandapat, and Sunayana Sitaram.
\newblock A case study of efficacy and challenges in practical human-in-loop
  evaluation of nlp systems using checklist.
\newblock In {\em Proceedings of the Workshop on Human Evaluation of NLP
  Systems (HumEval)}, pages 120--130, 2021.

\bibitem{bogdoll2022one}
Daniel Bogdoll, Stefani Guneshka, and J~Marius Z{\"o}llner.
\newblock One ontology to rule them all: Corner case scenarios for autonomous
  driving.
\newblock {\em arXiv preprint arXiv:2209.00342}, 2022.

\bibitem{gpt3}
Tom Brown, Benjamin Mann, Nick Ryder, Melanie Subbiah, Jared~D Kaplan, Prafulla
  Dhariwal, Arvind Neelakantan, Pranav Shyam, Girish Sastry, Amanda Askell,
  Sandhini Agarwal, Ariel Herbert-Voss, Gretchen Krueger, Tom Henighan, Rewon
  Child, Aditya Ramesh, Daniel Ziegler, Jeffrey Wu, Clemens Winter, Chris
  Hesse, Mark Chen, Eric Sigler, Mateusz Litwin, Scott Gray, Benjamin Chess,
  Jack Clark, Christopher Berner, Sam McCandlish, Alec Radford, Ilya Sutskever,
  and Dario Amodei.
\newblock Language models are few-shot learners.
\newblock In H. Larochelle, M. Ranzato, R. Hadsell, M.~F. Balcan, and H. Lin,
  editors, {\em Advances in Neural Information Processing Systems}, volume~33,
  pages 1877--1901. Curran Associates, Inc., 2020.

\bibitem{deng2009imagenet}
Jia Deng, Wei Dong, Richard Socher, Li-Jia Li, Kai Li, and Li Fei-Fei.
\newblock Imagenet: A large-scale hierarchical image database.
\newblock In {\em Conference on Computer Vision and Pattern Recognition
  (CVPR)}, 2009.
\newblock \url{https://ieeexplore.ieee.org/abstract/document/5206848}.

\bibitem{denton2019image}
Emily Denton, Ben Hutchinson, Margaret Mitchell, Timnit Gebru, and Andrew
  Zaldivar.
\newblock Image counterfactual sensitivity analysis for detecting unintended
  bias.
\newblock {\em arXiv preprint arXiv:1906.06439}, 2019.

\bibitem{d2022spotlight}
Greg d'Eon, Jason d'Eon, James~R Wright, and Kevin Leyton-Brown.
\newblock The spotlight: A general method for discovering systematic errors in
  deep learning models.
\newblock In {\em 2022 ACM Conference on Fairness, Accountability, and
  Transparency}, pages 1962--1981, 2022.

\bibitem{devlin-etal-2019-bert}
Jacob Devlin, Ming-Wei Chang, Kenton Lee, and Kristina Toutanova.
\newblock {BERT}: Pre-training of deep bidirectional transformers for language
  understanding.
\newblock In {\em Proceedings of the 2019 Conference of the North {A}merican
  Chapter of the Association for Computational Linguistics: Human Language
  Technologies, Volume 1 (Long and Short Papers)}, pages 4171--4186,
  Minneapolis, Minnesota, June 2019. Association for Computational Linguistics.

\bibitem{dosovitskiy2020image}
Alexey Dosovitskiy, Lucas Beyer, Alexander Kolesnikov, Dirk Weissenborn,
  Xiaohua Zhai, Thomas Unterthiner, Mostafa Dehghani, Matthias Minderer, Georg
  Heigold, Sylvain Gelly, et~al.
\newblock An image is worth 16x16 words: Transformers for image recognition at
  scale.
\newblock {\em arXiv preprint arXiv:2010.11929}, 2020.

\bibitem{du2022vision}
Xin Du, Benedicte Legastelois, Bhargavi Ganesh, Ajitha Rajan, Hana Chockler,
  Vaishak Belle, Stuart Anderson, and Subramanian Ramamoorthy.
\newblock Vision checklist: Towards testable error analysis of image models to
  help system designers interrogate model capabilities.
\newblock {\em arXiv preprint arXiv:2201.11674}, 2022.

\bibitem{eyuboglu2022domino}
Sabri Eyuboglu, Maya Varma, Khaled Saab, Jean-Benoit Delbrouck, Christopher
  Lee-Messer, Jared Dunnmon, James Zou, and Christopher R{\'e}.
\newblock Domino: Discovering systematic errors with cross-modal embeddings.
\newblock {\em arXiv preprint arXiv:2203.14960}, 2022.

\bibitem{ganguli2022red}
Deep Ganguli, Liane Lovitt, Jackson Kernion, Amanda Askell, Yuntao Bai, Saurav
  Kadavath, Ben Mann, Ethan Perez, Nicholas Schiefer, Kamal Ndousse, et~al.
\newblock Red teaming language models to reduce harms: Methods, scaling
  behaviors, and lessons learned.
\newblock {\em arXiv preprint arXiv:2209.07858}, 2022.

\bibitem{he2016deep}
Kaiming He, Xiangyu Zhang, Shaoqing Ren, and Jian Sun.
\newblock Deep residual learning for image recognition.
\newblock In {\em Proceedings of the IEEE conference on computer vision and
  pattern recognition}, pages 770--778, 2016.

\bibitem{imagenetr}
Dan Hendrycks, Steven Basart, Norman Mu, Saurav Kadavath, Frank Wang, Evan
  Dorundo, Rahul Desai, Tyler Zhu, Samyak Parajuli, Mike Guo, Dawn Song, Jacob
  Steinhardt, and Justin Gilmer.
\newblock The many faces of robustness: A critical analysis of
  out-of-distribution generalization.
\newblock {\em ICCV}, 2021.

\bibitem{imageneta}
Dan Hendrycks, Kevin Zhao, Steven Basart, Jacob Steinhardt, and Dawn Song.
\newblock Natural adversarial examples.
\newblock {\em CVPR}, 2021.

\bibitem{jain2022distilling}
Saachi Jain, Hannah Lawrence, Ankur Moitra, and Aleksander Madry.
\newblock Distilling model failures as directions in latent space.
\newblock {\em arXiv preprint arXiv:2206.14754}, 2022.

\bibitem{johnson2023does}
Nari Johnson, {\'A}ngel~Alexander Cabrera, Gregory Plumb, and Ameet Talwalkar.
\newblock Where does my model underperform? a human evaluation of slice
  discovery algorithms.
\newblock {\em arXiv preprint arXiv:2306.08167}, 2023.

\bibitem{khorram2022cycle}
Saeed Khorram and Li Fuxin.
\newblock Cycle-consistent counterfactuals by latent transformations.
\newblock In {\em Proceedings of the IEEE/CVF Conference on Computer Vision and
  Pattern Recognition}, pages 10203--10212, 2022.

\bibitem{kiela2021dynabench}
Douwe Kiela, Max Bartolo, Yixin Nie, Divyansh Kaushik, Atticus Geiger,
  Zhengxuan Wu, Bertie Vidgen, Grusha Prasad, Amanpreet Singh, Pratik Ringshia,
  et~al.
\newblock Dynabench: Rethinking benchmarking in nlp.
\newblock {\em arXiv preprint arXiv:2104.14337}, 2021.

\bibitem{negative_labels}
Youngdong Kim, Junho Yim, Juseung Yun, and Junmo Kim.
\newblock Nlnl: Negative learning for noisy labels.
\newblock In {\em Proceedings of the IEEE/CVF International Conference on
  Computer Vision}, pages 101--110, 2019.

\bibitem{leclerc20213db}
Guillaume Leclerc, Hadi Salman, Andrew Ilyas, Sai Vemprala, Logan Engstrom,
  Vibhav Vineet, Kai Xiao, Pengchuan Zhang, Shibani Santurkar, Greg Yang,
  et~al.
\newblock 3db: A framework for debugging computer vision models.
\newblock {\em arXiv preprint arXiv:2106.03805}, 2021.

\bibitem{loshchilov2018decoupled}
Ilya Loshchilov and Frank Hutter.
\newblock Decoupled weight decay regularization.
\newblock In {\em International Conference on Learning Representations (ICLR)},
  2019.
\newblock \url{https://openreview.net/forum?id=Bkg6RiCqY7}.

\bibitem{ma2018group}
Kede Ma, Zhengfang Duanmu, Zhou Wang, Qingbo Wu, Wentao Liu, Hongwei Yong,
  Hongliang Li, and Lei Zhang.
\newblock Group maximum differentiation competition: Model comparison with few
  samples.
\newblock {\em IEEE Transactions on pattern analysis and machine intelligence},
  42(4):851--864, 2018.

\bibitem{ma2022crepe}
Zixian Ma, Jerry Hong, Mustafa~Omer Gul, Mona Gandhi, Irena Gao, and Ranjay
  Krishna.
\newblock Crepe: Can vision-language foundation models reason compositionally?
\newblock {\em arXiv preprint arXiv:2212.07796}, 2022.

\bibitem{mitchell2019model}
Margaret Mitchell, Simone Wu, Andrew Zaldivar, Parker Barnes, Lucy Vasserman,
  Ben Hutchinson, Elena Spitzer, Inioluwa~Deborah Raji, and Timnit Gebru.
\newblock Model cards for model reporting.
\newblock In {\em Proceedings of the conference on fairness, accountability,
  and transparency}, pages 220--229, 2019.

\bibitem{paszke2019pytorch}
Adam Paszke, Sam Gross, Francisco Massa, Adam Lerer, James Bradbury, Gregory
  Chanan, Trevor Killeen, Zeming Lin, Natalia Gimelshein, Luca Antiga, et~al.
\newblock Pytorch: An imperative style, high-performance deep learning library.
\newblock In {\em Advances in Neural Information Processing Systems (NeurIPS)},
  2019.
\newblock \url{https://arxiv.org/abs/1912.01703}.

\bibitem{patel2008investigating}
Kayur Patel, James Fogarty, James~A Landay, and Beverly Harrison.
\newblock Investigating statistical machine learning as a tool for software
  development.
\newblock In {\em Proceedings of the SIGCHI Conference on Human Factors in
  Computing Systems}, pages 667--676. ACM, 2008.

\bibitem{radford2021learning}
Alec Radford, Jong~Wook Kim, Chris Hallacy, Aditya Ramesh, Gabriel Goh,
  Sandhini Agarwal, Girish Sastry, Amanda Askell, Pamela Mishkin, Jack Clark,
  et~al.
\newblock Learning transferable visual models from natural language
  supervision.
\newblock In {\em International Conference on Machine Learning}, pages
  8748--8763. PMLR, 2021.

\bibitem{ramasesh2021effect}
Vinay~Venkatesh Ramasesh, Aitor Lewkowycz, and Ethan Dyer.
\newblock Effect of scale on catastrophic forgetting in neural networks.
\newblock In {\em International Conference on Learning Representations}, 2021.

\bibitem{ramesh2022hierarchical}
Aditya Ramesh, Prafulla Dhariwal, Alex Nichol, Casey Chu, and Mark Chen.
\newblock Hierarchical text-conditional image generation with clip latents.
\newblock {\em arXiv preprint arXiv:2204.06125}, 2022.

\bibitem{imagenettoimagenet}
Benjamin Recht, Rebecca Roelofs, Ludwig Schmidt, and Vaishaal Shankar.
\newblock Do imagenet classifiers generalize to imagenet?
\newblock In {\em International Conference on Machine Learning}, pages
  5389--5400, 2019.

\bibitem{recht2019imagenet}
Benjamin Recht, Rebecca Roelofs, Ludwig Schmidt, and Vaishaal Shankar.
\newblock Do imagenet classifiers generalize to imagenet?
\newblock In {\em International Conference on Machine Learning}, pages
  5389--5400. PMLR, 2019.

\bibitem{ribeiro2022adaptive}
Marco~Tulio Ribeiro and Scott Lundberg.
\newblock Adaptive testing and debugging of nlp models.
\newblock In {\em Proceedings of the 60th Annual Meeting of the Association for
  Computational Linguistics (Volume 1: Long Papers)}, pages 3253--3267, 2022.

\bibitem{ribeiro2020beyond}
Marco~Tulio Ribeiro, Tongshuang Wu, Carlos Guestrin, and Sameer Singh.
\newblock Beyond accuracy: Behavioral testing of nlp models with checklist.
\newblock In {\em Association for Computational Linguistics (ACL)}, 2020.

\bibitem{saharia2022photorealistic}
Chitwan Saharia, William Chan, Saurabh Saxena, Lala Li, Jay Whang, Emily
  Denton, Seyed Kamyar~Seyed Ghasemipour, Burcu~Karagol Ayan, S~Sara Mahdavi,
  Rapha~Gontijo Lopes, et~al.
\newblock Photorealistic text-to-image diffusion models with deep language
  understanding.
\newblock {\em arXiv preprint arXiv:2205.11487}, 2022.

\bibitem{schuhmann2022laion}
Christoph Schuhmann, Romain Beaumont, Richard Vencu, Cade Gordon, Ross
  Wightman, Mehdi Cherti, Theo Coombes, Aarush Katta, Clayton Mullis, Mitchell
  Wortsman, et~al.
\newblock Laion-5b: An open large-scale dataset for training next generation
  image-text models.
\newblock {\em arXiv preprint arXiv:2210.08402}, 2022.

\bibitem{sohoni2020no}
Nimit Sohoni, Jared Dunnmon, Geoffrey Angus, Albert Gu, and Christopher R{\'e}.
\newblock No subclass left behind: Fine-grained robustness in coarse-grained
  classification problems.
\newblock {\em Advances in Neural Information Processing Systems},
  33:19339--19352, 2020.

\bibitem{tian2018deeptest}
Yuchi Tian, Kexin Pei, Suman Jana, and Baishakhi Ray.
\newblock Deeptest: Automated testing of deep-neural-network-driven autonomous
  cars.
\newblock In {\em Proceedings of the 40th international conference on software
  engineering}, pages 303--314, 2018.

\bibitem{wang2020going}
Haotao Wang, Tianlong Chen, Zhangyang Wang, and Kede Ma.
\newblock I am going mad: Maximum discrepancy competition for comparing
  classifiers adaptively.
\newblock {\em arXiv preprint arXiv:2002.10648}, 2020.

\bibitem{imagenetsketch}
Haohan Wang, Songwei Ge, Zachary Lipton, and Eric~P Xing.
\newblock Learning robust global representations by penalizing local predictive
  power.
\newblock In {\em Advances in Neural Information Processing Systems}, pages
  10506--10518, 2019.

\bibitem{wang2022ofa}
Peng Wang, An Yang, Rui Men, Junyang Lin, Shuai Bai, Zhikang Li, Jianxin Ma,
  Chang Zhou, Jingren Zhou, and Hongxia Yang.
\newblock Ofa: Unifying architectures, tasks, and modalities through a simple
  sequence-to-sequence learning framework.
\newblock {\em CoRR}, abs/2202.03052, 2022.

\bibitem{wang2022unifying}
Peng Wang, An Yang, Rui Men, Junyang Lin, Shuai Bai, Zhikang Li, Jianxin Ma,
  Chang Zhou, Jingren Zhou, and Hongxia Yang.
\newblock Unifying architectures, tasks, and modalities through a simple
  sequence-to-sequence learning framework.
\newblock {\em arXiv preprint arXiv:2202.03052}, 2022.

\bibitem{wang2008maximum}
Zhou Wang and Eero~P Simoncelli.
\newblock Maximum differentiation (mad) competition: A methodology for
  comparing computational models of perceptual quantities.
\newblock {\em Journal of Vision}, 8(12):8--8, 2008.

\bibitem{wiles2022discovering}
Olivia Wiles, Isabela Albuquerque, and Sven Gowal.
\newblock Discovering bugs in vision models using off-the-shelf image
  generation and captioning.
\newblock {\em arXiv preprint arXiv:2208.08831}, 2022.

\bibitem{wortsman2022robust}
Mitchell Wortsman, Gabriel Ilharco, Jong~Wook Kim, Mike Li, Simon Kornblith,
  Rebecca Roelofs, Raphael~Gontijo Lopes, Hannaneh Hajishirzi, Ali Farhadi,
  Hongseok Namkoong, et~al.
\newblock Robust fine-tuning of zero-shot models.
\newblock In {\em Proceedings of the IEEE/CVF Conference on Computer Vision and
  Pattern Recognition}, pages 7959--7971, 2022.

\bibitem{yan2021exposing}
Jiebin Yan, Yu Zhong, Yuming Fang, Zhangyang Wang, and Kede Ma.
\newblock Exposing semantic segmentation failures via maximum discrepancy
  competition.
\newblock {\em International Journal of Computer Vision}, 129:1768--1786, 2021.

\bibitem{yuksekgonul2022and}
Mert Yuksekgonul, Federico Bianchi, Pratyusha Kalluri, Dan Jurafsky, and James
  Zou.
\newblock When and why vision-language models behave like bag-of-words models,
  and what to do about it?
\newblock {\em arXiv preprint arXiv:2210.01936}, 2022.

\bibitem{zendel2018wilddash}
Oliver Zendel, Katrin Honauer, Markus Murschitz, Daniel Steininger, and
  Gustavo~Fernandez Dominguez.
\newblock Wilddash-creating hazard-aware benchmarks.
\newblock In {\em Proceedings of the European Conference on Computer Vision
  (ECCV)}, pages 402--416, 2018.

\bibitem{zhang2018deeproad}
Mengshi Zhang, Yuqun Zhang, Lingming Zhang, Cong Liu, and Sarfraz Khurshid.
\newblock Deeproad: Gan-based metamorphic testing and input validation
  framework for autonomous driving systems.
\newblock In {\em 2018 33rd IEEE/ACM International Conference on Automated
  Software Engineering (ASE)}, pages 132--142. IEEE, 2018.

\bibitem{zhao2021calibrate}
Zihao Zhao, Eric Wallace, Shi Feng, Dan Klein, and Sameer Singh.
\newblock Calibrate before use: Improving few-shot performance of language
  models.
\newblock In {\em International Conference on Machine Learning}, pages
  12697--12706. PMLR, 2021.

\end{thebibliography}
}

\clearpage
\appendix
\section*{Appendix Overview}
\begin{itemize}
    \item In Appendix \ref{app:system}, we present additional details on the \tool~system as described in Section 3, including details about adaptive test retrieval and automatic test labeling (\ref{app:inl}), as well as adaptive topic generation based on templates and user topics (\ref{app:outl}).
    \item In Appendix \ref{app:user_studies}, we include additional details about the user studies in Section 4.2, as well as details about the statistical analyses.
    \item In Appendix \ref{app:domino}, we expand upon the comparison of \tool~with \domino~(Section 4.3 in the main text), providing an explanation of \domino~and its hyperparameters, our criterion for coherency, a list of all slice descriptions (topics) from \domino~used during evaluation, and a list of the 30 user-found \tool~topics we compared \domino~against.
    \item Finally, in Appendix \ref{app:finetuning}, we discuss hyperparameters used for finetuning in Section 4.4, list all control and treatment topics, and include an additional evaluation measuring whether finetuning on treatment topics improves other, conceptually unrelated bugs.
\end{itemize}

\begin{algorithm*}[bth]
    \KwIn{
        Textual topic description $z$, previously labeled tests $\mathcal D = \{(x, \model(x), y)\}$, previous off-topic tests $\mathcal D_\text{off-topic}$
    } 
    \vspace{0.5em}
    Compute $q_t \leftarrow \text{\clip}(z)$\Comment{Figure 2A}\\
    \eIf{$|\mathcal D| > 0$}{ 
        Sample $x_1, x_2, x_3 \sim \text{Categorical}(|\mathcal D|, p_j)$, where $p_j$ is computed according to the text in \ref{app:inl}\\
        Aggregate $q_i \leftarrow \sum_k \beta_k \cdot \text{\clip}(x_k)$, with $\beta \sim \text{Dirichlet}(1,1,1)$\\
        Set $q \leftarrow \text{slerp}(q_t, q_i, r)$, with $r \sim \text{Uni}(0,1)$\\
    }
    {
        Set $q \leftarrow q_t$
    }
    \vspace{0.5em}
    Retrieve approximate nearest neighbors of $q$ from \laion~\Comment{Figure 2B}\\
    Exclude retrievals whose \clip~image embeddings have cosine similarity $> 0.9$ with any previous test $x \in \mathcal D$\\
    Collect model outputs for all retrieved images to obtain new collection of tests $\mathcal S \leftarrow [(\tilde x, \model(\tilde x))]$\\
    \vspace{0.5em}
    \If{$|\mathcal D| > 0$}{ \Comment{Figure 2C}\\
        Train a lightweight classifier $f$ on previously labeled tests $\mathcal D$ as described in \ref{app:inl}\\
        Sort $\mathcal S$ according to $f(\tilde x)$ for $\tilde x \in \mathcal S$, placing predicted fails far from the decision boundary first, and predicted passes far from the decision boundary last
        Update $\mathcal S$ to contain $(\tilde x, \model(\tilde x), f(x))$, so that we can display the imputed label to the user\\
        Train a second lightweight classifier $f_\text{off-topic}$ to differentiate between previous in-topic tests $\mathcal D$ and previous off-topic tests $\mathcal D_\text{off-topic}$\\
        Place tests $\tilde x \in \mathcal S$ for which $f_\text{off-topic}(x)$ predicts ``off-topic'' at the end of $\mathcal S$\\ 
    }
    \Return sorted $\mathcal S$ to the user for confirmation / correction. \Comment{Figure 2D}
    \caption{Iteration of the \inl.}
    \label{alg:inl}
\end{algorithm*}

\section{Additional details on \tool}\label{app:system}
In Section 3, we described \tool, which includes a \inl~that retrieves images with \clip~and an \outl~that generates topics with \gpt.
Here, we expand on the details of both loops.

\subsection{\Inl}\label{app:inl}
In the \inl, users explore a candidate topic $t$.
Each iteration of the loop, \tool~retrieves test suggestions relevant to the topic, and, when possible, automatically imputes pass/fail labels for these suggestions in order to minimize user labeling effort.
In this section, we provide additional details about the retrieval and labeling steps of a single iteration of the \inl.

Recall that $\model$ is our target vision model (\eg a classification model), and $\model(x)$ is the model output on an image $x$ (\eg the label string, such as ``banana'').
At any given iteration of the \inl, we have a textual topic description $z$ and a (possibly empty) set of already labeled tests $\mathcal D$, where each test is a triplet $(x, \model(x), y)$. 
The label $y \in \{-1, 1\}$ refers to whether the test failed (-1) or passed (1).
We may also have a set of previously reviewed, off-topic tests $\mathcal D_\text{off-topic}$.
In Algorithm \ref{alg:inl}, we step through the retrieval and labeling steps of the iteration.
For space, we expand on two of the steps in the algorithm box below.


\paragraph{Sampling previous tests $x_1, x_2, x_3$ for retrieval.} In the retrieval step, we incorporate three previous tests from $\mathcal D$.
Each of the three tests is sampled from a categorical distribution over $\mathcal D$, where each $x_j \in \mathcal D$ has probability $p_j$ of being selected.
We prefer to select tests where the model \textit{confidently} fails over tests where the model less confidently fails, which are still preferred over passed tests.
Thus, when available, we use the model's \textit{prediction confidence} $s_\model(x_j) \in [0, 1]$ for each example $x_j \in \mathcal D$ to compute $p_j$.
For example, in classification, $s_\model$ is the model's maximum softmax score.
If this value is unavailable (\eg~for some commercial APIs), we fix $s_\model(x_j) = 1$ for all examples. 
To compute $p_j$, we compute a score $\alpha_j = 1 - y_j s_\model(x_j)$ per example (note that this is $0$ when the test is confidently correct and $2$ when the model is confidently incorrect), and we set $p_j$ to be the normalized version of $\alpha_j$. 

\paragraph{Lightweight classifiers $f, f_\text{off-topic}$ for automatically labeling tests.} 
When $|\mathcal D| > 0$, we use two lightweight classifiers to automatically impute whether tests have passed, failed, or are off-topic. 
The lightweight classifiers are specialized to the current topic (\ie we train new classifiers for new topics).
Functionally, a lightweight classifier is a Support Vector Classifiers (SVC).
One lightweight classifier $f$ maps $(x, \model(x))$ to predicted pass/fail label in $\{-1, 1\}$. 
Specifically, the tuple $(x, \model(x))$ is transformed into a single vector by first embedding both $x$ (image) and $\model(x)$ (string) with \clipvit, followed by concatenating the two into $[\text{\clip}(x), \text{\clip}(\model(x))]$.
This vector is used as the feature representation of the SVC. 
New tests are re-sorted according to the prediction of $f$ and $f$'s confidence, with likely failures shown first. 
Classifier $f_\text{off-topic}$ operates on the same representation as $f$ for each test, but instead predicts binary labels for in-topic / off-topic.
These classifiers take less than a second to train and run, so we re-train them at each iteration of the \inl.

\subsection{\Outl}\label{app:outl}
In the \outl, users collaborate with \tool~to generate candidate topics to explore using \gpt.
We generate candidate topics in two phases. 
In the first phase, we prompt \gpt~using a pre-written set of templates and collect the completions.
We share the set of prompt templates used in Listing \ref{lst:prompts}, replacing \{LABEL\} with predefined label names or existing user topics (\eg \topic{stop sign}).
In the second phase, we gather suggestions from the first round, append previously explored topics with high failure rates, and gather a second round of suggestions.
We place topics with the highest failure rates at the end of the prompts to account for \gpt's recency bias~\cite{zhao2021calibrate}.
Finally, topics are presented to users, who explore ones they deem interesting and important.

\begin{lstlisting}[label=lst:prompts,caption=Example prompts used in \outl.,float,frame=tb]
List some unexpected places to see a {LABEL}
List some places to find a {LABEL}
List some other things that you usually find
    with a {LABEL}
List some artistic representations of a {LABEL}
List some things that can be made to look like 
    a {LABEL}
List some types of {LABEL} you wouldn't normally
    see
List some dramatic conditions to photograph
    a {LABEL}
List some conditions a {LABEL} could be in that 
    would make it hard to see
List some things that are the same shape as a
    {LABEL}
List some {LABEL} that are a different color than 
    you would expect
\end{lstlisting}

\subsection{Web interface}
\begin{figure*}
\centering
\begin{subfigure}[b]{0.52\textwidth}
    \includegraphics[width=\textwidth]{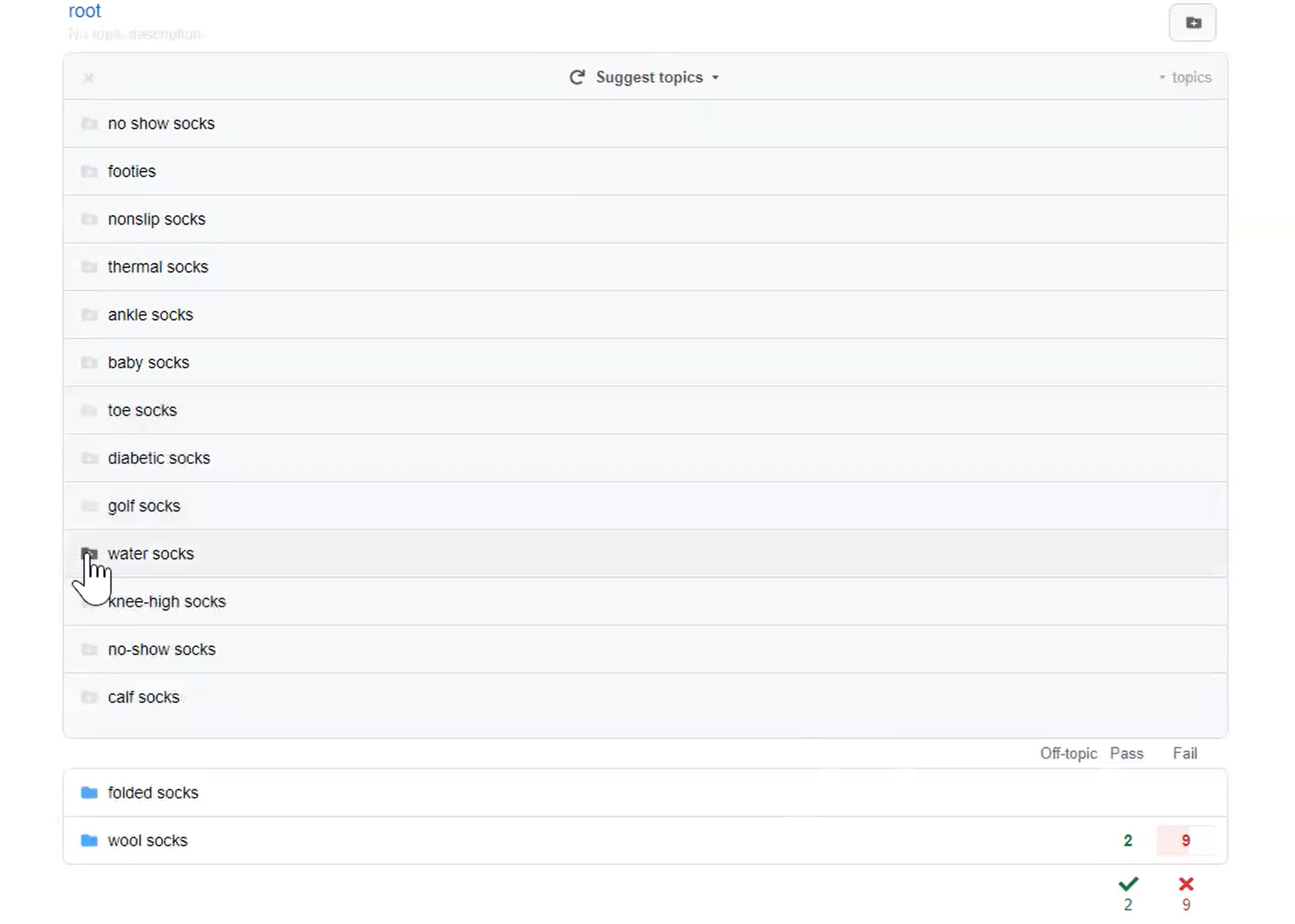}
\end{subfigure}
\hfill
\begin{subfigure}[b]{0.47\textwidth}
    \includegraphics[width=\textwidth]{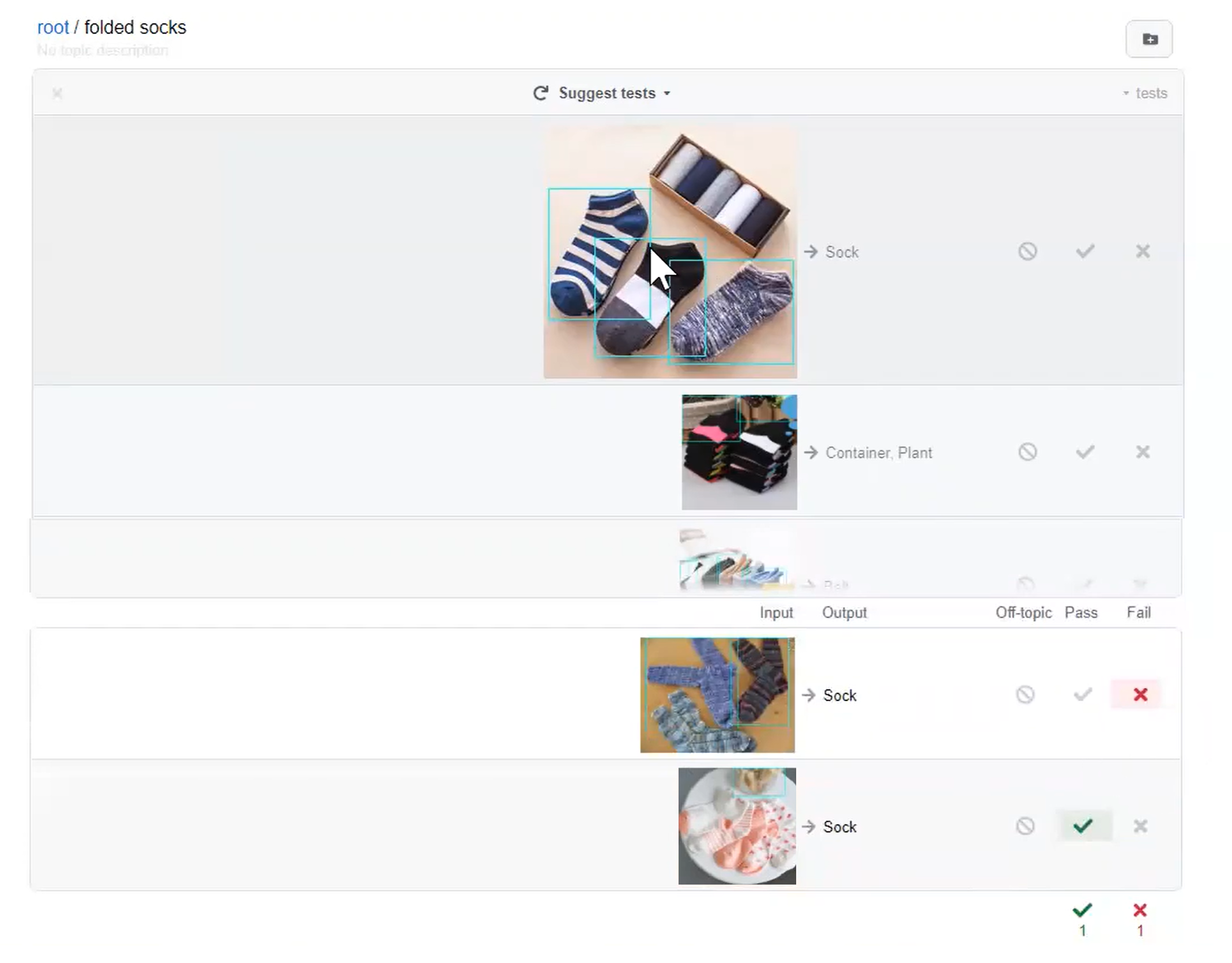}
\end{subfigure}
\caption{In the \tool~web interface, topics are represented as folders, and tests are represented as rows mapping images to model outputs.
The \outl is represented as a root page that suggests topics to explore (left), while the \inl within each topic suggests tests that lightweight classifiers label as potential failures (right, top panel).
}
\label{fig:web_interface}
\end{figure*}

We provide screenshots of the \tool~web interface in Figure \ref{fig:web_interface}.
The \outl is represented as a root page that suggests topics to explore, and individual topics are represented as folders (Figure \ref{fig:web_interface} left).
Tests within folders are represented as rows mapping images to model outputs (Figure \ref{fig:web_interface} right).

\section{Additional details for user studies}\label{app:user_studies}
In Section 4.2, we described a large set of user studies used to evaluate \tool's ability to enable users to find bugs in state-of-the-art vision models. Here, we include additional details about the study setups and statistical analyses.

\paragraph{Study setup.}
All participants undertook the study virtually in a single 60-minute Zoom session.
At the start of the session, participants were shown a 5-minute video introducing how to use the \tool web interface.
Next, the experimenter walked through instructions for the testing task:
as described in the main text, participants were instructed to find as many failure-prone topics (bugs) for a specific category (\eg banana) as possible, and to switch topics whenever they found more than a threshold of failures within a topic.
This latter instruction prevented users from endlessly exploiting a topic to inflate the total failure count.
We adjusted the threshold at which a topic became a bug (and users should move on) based on the time users had for testing: for \class~and \capt, users tested models for 20 minutes with a bug threshold of 10 failures, while for \detc, users tested the model for 15 minutes with a bug threshold of 8.

\begin{figure*}
    \centering
    \includegraphics[height=8.5in]{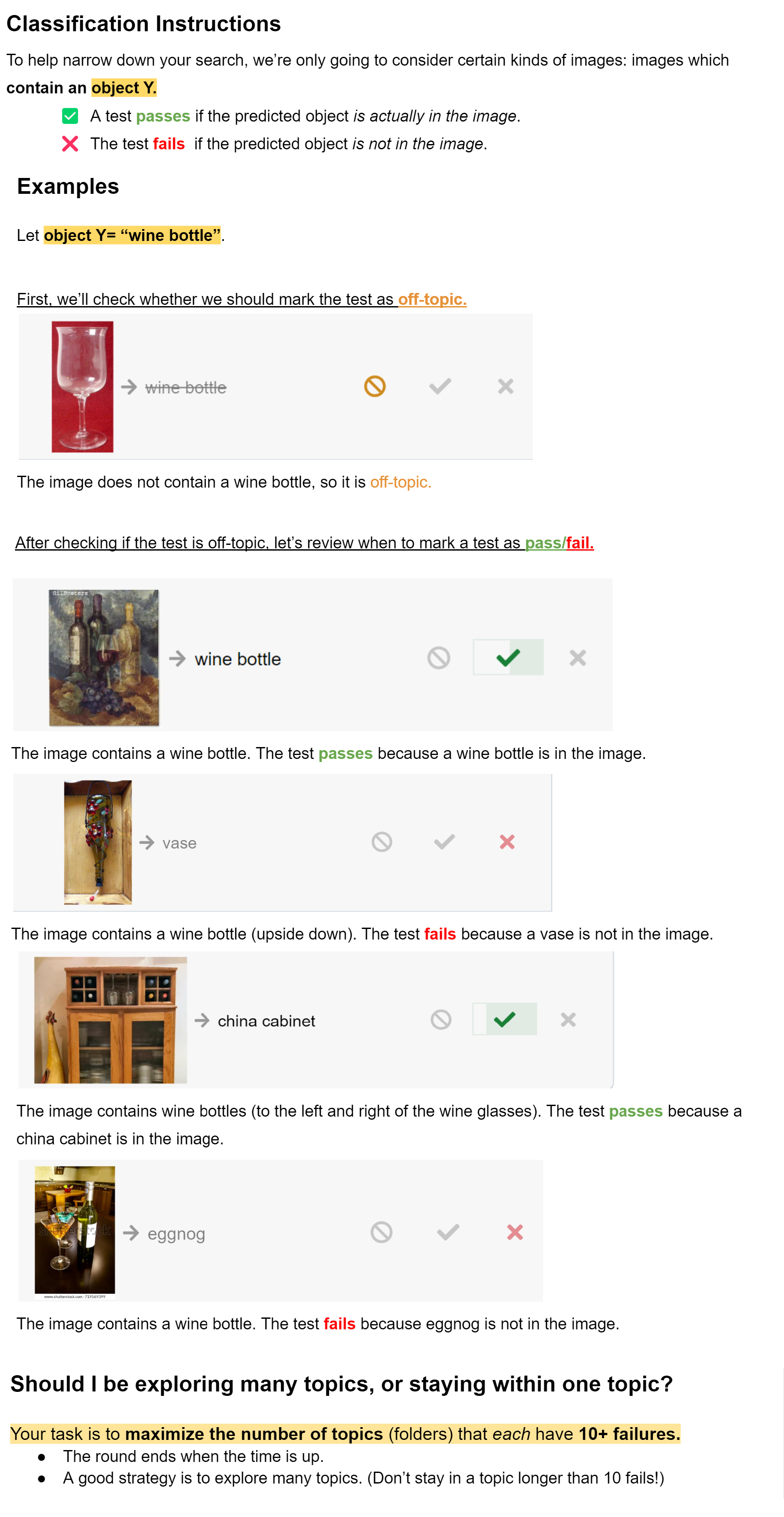}
    \caption{
        Example instructions for \class~users.
        \label{fig:class_instructions}
    }
\end{figure*}

\begin{figure*}
    \centering
    \includegraphics[height=8.5in]{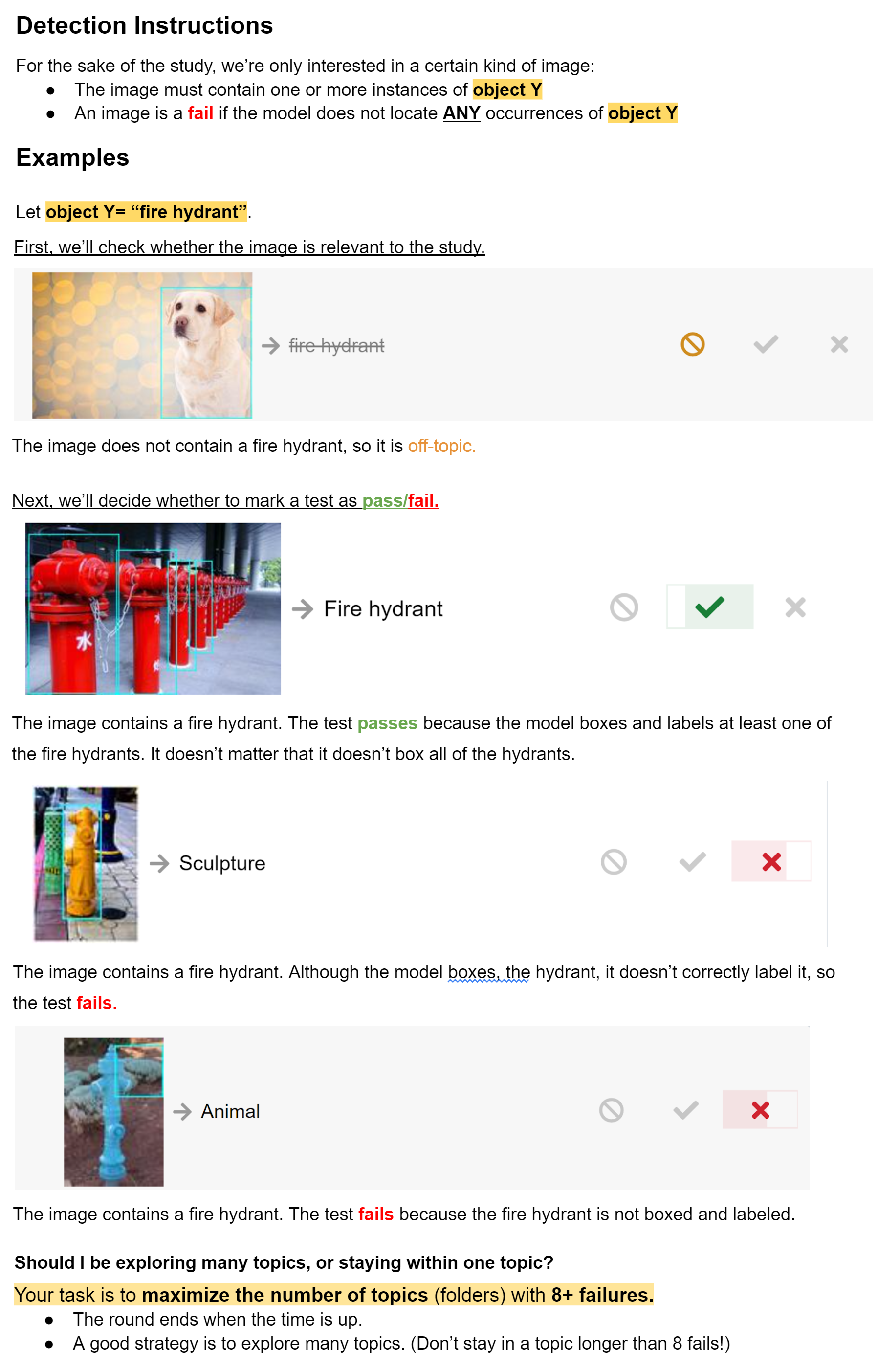}
    \caption{
        Example instructions for \detc~users.
        \label{fig:detc_instructions}
    }
\end{figure*}

\begin{figure*}
    \centering
    \includegraphics[height=8.5in]{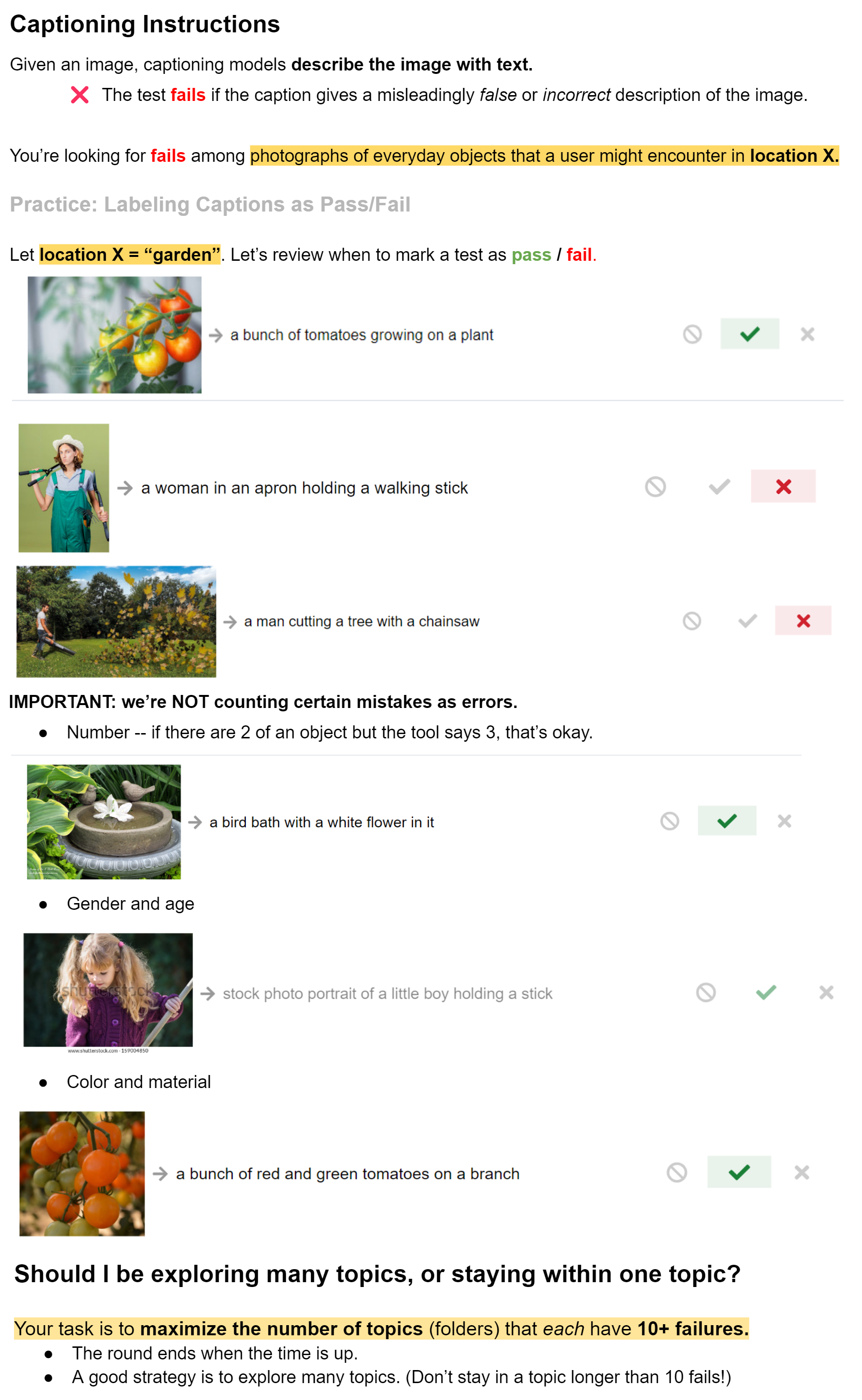}
    \caption{
        Example instructions for \capt~users.
        \label{fig:capt_instructions}
    }
\end{figure*}

When introducing the task, the experimenter defined failed tests as follows:

\begin{itemize}
\item For participants testing \class~models, a test failed if the model predicted an object not present in the image. Users were instructed to look for failures among pictures of a specific category (\textit{banana} or \textit{broom}), \eg failures among pictures of bananas (Figure \ref{fig:class_instructions}). Participants were given class definitions from an ImageNet labeling guide used in \cite{imagenettoimagenet}.
\item For those testing \detc~models, a test failed if the model failed to box any instance of the given category (\textit{bicycle} or \textit{stop sign}), \eg as shown in Figure \ref{fig:detc_instructions}.
\item Participants testing \capt~models were asked to imagine that they were testing a product used by visually impaired customers to caption everyday scenes. The participants' task was to find images (tests) for which the model produced false or incorrect captions, excluding counting, color, and gender or age mistakes (Figure \ref{fig:capt_instructions}). Participants looked for such tests among pictures of a specific category (\ie scenes customers might encounter in a \textit{kitchen} or \textit{elementary school}).
\end{itemize}

The experimenter then asked each user to practice using the web interface by testing the model a third, held-out object or location for 10 minutes.
For \class, this was a wine bottle; for \detc, this was a fire hydrant; and for \capt, this was a \textit{garden}.
After two rounds of testing in the main experiment, the study concluded with an exit survey as described in Table \ref{tab:exit_survey}.
The study compensation was a \$25 Amazon gift card.

\begin{table}[ht]
    \centering
    \begin{tabular}{|p{0.4\linewidth}|p{0.4\linewidth}|}
    \hline
    \textbf{Question} & \textbf{Type} \\
    \hline
    How difficult was it to find bugs in the first round? & 5-point Likert scale \\
    \hline
    How difficult was it to find bugs in the second round? & 5-point Likert scale \\
    \hline
    How useful was the web tool for finding bugs? & Multiple-choice \{I could have found these bugs using existing error analysis tools I have access to, I could not have found these bugs using existing error analysis tools I have access to.\} \\
    \hline
    Did you use topic suggestions? Were they helpful? & Multiple-choice \{Yes, and they were helpful in generating topics that caused failures, Yes, and they were helpful in generating ideas for topics to explore, Yes, but they did not generate good ideas for topics to explore, No, I did not use topic suggestions.\}\\
    \hline
    \end{tabular}
    \caption{Items in exit survey. \label{tab:exit_survey}}
\end{table}

\paragraph{Additional results.}
As discussed in Section 4.2, \tool~helped users find significantly more failing tests than \baseline, with significance determined by paired t-tests in each task.
In \class, $t(16)=2.27, p<0.05$; in \detc, $t(16)=3.42, p<0.005$, and in \capt, $t(8)=2.56, p<0.05$.
These corresponded to normalized effect sizes of $d=0.588$ in \class, $d=0.882$ in \detc, and $d=0.967$ in \capt.
We also counted the number of users who could find bugs during testing, \ie identify a topic that hit the threshold number of fails (8 for \detc, 10 for the other tasks).
Overall, 28/40 users found bugs during testing with \tool, while only 16/40 could find such a high-failure topics in the baseline round. 
When surveyed about perceived difficulty of finding bugs, users also felt finding bugs was easier with \tool~than without, with $t(40)=4.18, p < 0.0005$.
When surveyed about the helpfulness of \gpt's topic suggestions, 24/34 users who used the topic suggestions marked that they were helpful for exploration.

\section{Additional details for comparison with automatic slice discovery}\label{app:domino}
In Section 4.3, we compared \tool~with \domino~\cite{eyuboglu2022domino}, showing that bugs found with \tool~are more difficult.
Here, we provide an explanation of the \domino~method and its hyperparameters, define our criterion for coherency, list \textit{all} slice descriptions (topics) from \domino~used during evaluation (along with whether they satisfied coherency), and list the 30 user-found \tool~topics.

\paragraph{Details on \domino.}
We use the official release of \domino~\cite{eyuboglu2022domino}, available at \url{https://github.com/HazyResearch/domino}.
To generate slice proposals for a label in \sixoverlap, we ran the target model (\vit~or \resnet) over ImageNet validation examples of that class.
Then, we used \domino's error-aware mixture model to generate $5$ slices (per class).
The error-aware mixture model first generates $\bar k$ candidate slices (clusters of images) before selecting the best $5$ slices.
As in the original paper, we set $\bar k = 25$, and we initialize groups using the confusion matrix setting.
Additionally, \domino~uses a hyperparameter $\gamma = 10$ to control the weight placed on \textit{incorrect} examples when slicing.
In the original paper, the authors used $\gamma = 10$ for all datasets, which we also initially tried. 
However, we found that this setting produced slices that were too easy (no errors within the slices).
Thus, we matched the authors' blog post applying \domino~to ImageNet\footnote{\href{https://hazyresearch.stanford.edu/blog/2022-04-02-domino}{https://hazyresearch.stanford.edu/blog/2022-04-02-domino}}, conducting final experiments with $\gamma = 40$. 
(Larger values of $\gamma$ resulted in too much cluster incoherency.)

In our experiments, we evaluated two variations of \domino, which use the same image clusters but differ in their captioning strategy. 
The first, \dominobert, is the original proposal in~\cite{eyuboglu2022domino}; to caption a cluster, \dominobert~samples 250,000 BERT or Wikipedia completions, per cluster, of a set of templated captions; we provide the templates we used in Listing \ref{lst:domino}.
Each cluster is then matched to the caption with the highest cosine similarity in CLIP space.
We note that in the original paper, the authors used only one template: ``a photo of a [MASK]''; our modifications to this template account for the fact that all clusters are about a certain class (\eg clusters of banana images). Thus, we enforce that the label (banana) appears in the caption by populating \{LABEL\} in the template with the appropriate class name.
We also compared against our own variation of \domino, \dominoofa, which uses Alibaba's OFA-huge to more coherently caption clusters.
Here, we run \ofa~over all examples in a cluster and select the individual caption that maximizes cosine similarity with the cluster mean.

\begin{lstlisting}[label=lst:domino,caption=Templates used to generate captions for \dominobert.,float,frame=tb]
    a photo of {LABEL} and [MASK]
    a photo of {LABEL} in [MASK]
    a photo of [MASK] {LABEL}
    [MASK] {LABEL} [MASK]
\end{lstlisting}    

\paragraph{Coherency.} 
When reporting failure rates for \domino~topics, we calculated failure rates only over descriptions that were coherent. 
Descriptions were deemed incoherent if they were nonsensical (e.g. ``a photo of setup by banana'', ``a photo of skiing at sandal'') or did not refer to the target category at all (e.g. ``three oranges and an apple on a white background'' or ``a photo of promoter david lemon'' when the target is ``lemon''). 
For reference, we include the full list of \dominobert~descriptions for \vit~in Listing \ref{lst:dominobert-vit}, \resnet~in Listing \ref{lst:dominobert-resnet}, and the list of \dominoofa~descriptions for \vit~in Listing \ref{lst:dominoofa-vit} and \resnet~in Listing \ref{lst:dominoofa-resnet}.
In these lists, we prepend an asterisk in front of coherent topics, and we append the target category for each slice in parentheses.
Of the starred coherent descriptions, we excluded three from evaluation because we could not find any related images in \laion: ``a photo of munitions and broom'' (\dominobert, \vit), ``a photo of primate and broom'' (\dominobert, \resnet), and ``a large banana sitting in the middle of an abandoned building'' (\dominoofa, both \vit~and \resnet).

\paragraph{\tool~topics.} 
We also include the \tool~topics we compare to in Listing \ref{lst:adatest-topics}. 
Of the six categories \sixoverlap, we recruited one user per category to test \vit~and generate topics, except for the categories \textit{candle} and \textit{wine bottle}, which one of the authors tested in a separate session. 
All users (including the author) were limited to 20 minutes for testing, and they had not explored \vit~on the tested class before.


\section{Additional details for finetuning}
In Section 4.4, we presented results from experiments that show we can patch model performance on bugs while maintaining or slightly improving accuracy on the original ImageNet distribution, control topics, and OOD evaluation sets.
In these experiments, we finetuned on 20 images from each of 30 buggy topics (which we call the \textit{treatment topics}).
These treatment topics are the same as in Section 4.3 / Appendix \ref{app:domino}, listed in Listing \ref{lst:adatest-topics}.
In this section, we discuss hyperparameter choices, list all control topics, and break down OOD evaluation set gains by dataset.

\label{app:finetuning}
\begin{lstlisting}[label=lst:adatest-topics,caption=Topics from \tool~studied in Sections 4.3 and 4.4 in the main text.,float,frame=htb]
banana next to a banana smoothie
banana on kitchen countertop
banana in wooden woven basket
banana next to banana bread
toy banana
broom by fireplace
witch flying on a broom
photo of a person holding a boxy broom
silhouette of a person flying on a broom
broom in closet
black-and-white clipart of a candle
creamy white candle in glass jar
christmas candle next to tea
candle by window in snowstorm
person holding a candle at a vigil
grating a lemon
cooking with lemon
lemon tea with lemon
lemon on pancake with condensed milk or honey
lemon clipart
a lot of toy sandals
flip flop door wreath
translucent sandals
colorful flip flops
sandal ornament in a tree
top of a champagne bottle
wine bottle in a wiry wine rack
champagne in a champagne holder
wine bottle in a suit case
wine bottle with a wine stopper
\end{lstlisting}  

\begin{lstlisting}[label=lst:control,caption={Control topics in Section 4.4 in the main text. In parentheses, we list the topic found by \tool~with which the given control topic contrasts.},float,frame=tb]
shopping basket (banana in wooden woven basket)
bread (banana next to banana bread)
dishwasher in kitchen (banana on kitchen 
    countertop)
microwave in kitchen (banana on kitchen 
    countertop)
eggnog (banana next to a banana smoothie, 
    lemon on pancake with condensed milk or 
    honey)
witch with cauldron (witch flying on a broom)
fireplace no broom (broom by fireplace)
mop (broom in closet, 
    photo of a person holding a boxy broom)
person holding mop (photo of a person holding 
    a boxy broom)
consomme (lemon tea with lemon)
black tea no lemon (lemon tea with lemon)
grated orange (grating a lemon)
pancake with condensed milk or honey no lemon 
    (lemon on pancake with condensed milk or 
    honey)
torch (person holding a candle at a vigil,
    candle by window in snowstorm)
clog shoe (sandal)
beer bottle (top of a champagne bottle)
suit case (wine bottle in a suit case)
empty wine rack (wine bottle in a wiry wine 
    rack)
empty wire rack (wine bottle in a wiry wine 
    rack)
\end{lstlisting}  

\paragraph{Finetuning hyperparameters.}
We finetune \vit~using a small, constant learning rate of 1e-5 with the AdamW optimizer \cite{loshchilov2018decoupled, paszke2019pytorch} for five steps, with weight decay 0.01, batch size 16, and random square cropping for data augmentation. These hyperparameters were chosen in early experiments because they only degrade in-distribution model performance slightly. We report all results averaged over 3 random seeds along with the corresponding standard deviations.
When deduplicating evaluation data against the finetuning data, we mark pairs as duplicates if their CLIP cosine similarly $>0.95$.

\paragraph{Control topics.}
We provide a list of the 19 control topics from Section 4.4 in Listing \ref{lst:control}. 
These topics were selected because they were semantically related to the treatment topics in Listing \ref{lst:adatest-topics}, but had different labels.
For example, the \tool~topic \topic{person holding a boxy broom} is visually similar to the concept of ``person holding a mop'', so we include the latter as a control topic.
Other control topics are classes that were incorrectly predicted for a topic (\eg, \topic{banana on kitchen countertop} is frequently predicted ``microwave'', so we include ``microwave in kitchen'' as a control topic).
We checked performance on these topics to make sure performance gains were not due to the model forming new shortcuts. 

\paragraph{Per-OOD evaluation set breakdown.}
In Tables \ref{tab:ft-breakdown-treatment} and \ref{tab:ft-breakdown-global}, we provide a breakdown of the OOD evaluation set performances; these were aggregated as an average in Table 2 of the main text.
Table \ref{tab:ft-breakdown-treatment} displays the accuracy on treatment classes in each of the OOD evaluation sets, and Table \ref{tab:ft-breakdown-global} displays the overall accuracy in each of the OOD sets.

\begin{table*}
    \setlength\tabcolsep{3.5pt}
    \renewcommand{\arraystretch}{0.9}
    \small
    \begin{center}
    \resizebox{\textwidth}{!}{%
    \begin{tabular}{lccccccc}
    \toprule
    \textbf{Model} & \textbf{ImageNet} & \textbf{ImageNet V2} & \textbf{ImageNet-Sketch} & \textbf{ImageNet-R} & \textbf{ImageNet-A} & \textbf{ObjectNet} & \textbf{Avg. OOD}\\\midrule
    Before finetuning & 87.7 & 65.0 & 86.1 & 89.2 & 65.6 & 84.3 & 78.0 \\
    Finetuning with baseline & \textbf{93.1 (0.2)} & \textbf{71.7 (1.4)} & \textbf{90.9 (0.3)} & 90.5 (0.3) & 71.2 (1.3) & 86.4 (0.1) & 82.1 (0.6) \\
    Finetuning with \tool & \textbf{92.9 (0.4)} & \textbf{69.4 (0.8)} & \textbf{91.7 (0.5)} & \textbf{93.9 (0.2)} & \textbf{76.8 (1.5)} & \textbf{87.9 (0.2)} & \textbf{84.0 (0.2)} \\
    \bottomrule
    \end{tabular}
    }
    \end{center}
    \caption{
    Accuracies on treatment classes, before and after finetuning. Results are averaged over three random seeds.
    \label{tab:ft-breakdown-treatment}}
\end{table*}
    
\begin{table*}
    \setlength\tabcolsep{3.5pt}
    \renewcommand{\arraystretch}{0.9}
    \small
    \begin{center}
    \resizebox{\textwidth}{!}{%
    \begin{tabular}{lccccccc}
    \toprule
    \textbf{Model} & \textbf{ImageNet} & \textbf{ImageNet V2} & \textbf{ImageNet-Sketch} & \textbf{ImageNet-R} & \textbf{ImageNet-A} & \textbf{ObjectNet} & \textbf{Avg. OOD}\\\midrule
    Before finetuning & 88.4 & 81.0 & 64.4 & 89.1 & 83.9 & 69.9 & 77.7 \\
    Finetuning with baseline & \textbf{88.5 (0.0)} & \textbf{81.3 (0.0)} & 64.5 (0.0) & 89.2 (0.1) & 84.4 (0.1) & \textbf{70.5 (0.2)} & 78.0 (0.1) \\
    Finetuning with \tool & 88.4 (0.0) & 80.9 (0.1) & \textbf{64.7 (0.0)} & \textbf{90.0 (0.0)} &\textbf{84.9 (0.0)} & \textbf{70.5 (0.0)} & \textbf{78.2 (0.0)} \\
    \bottomrule
    \end{tabular}
    }
    \end{center}
    \caption{
    Accuracies on all classes, before and after finetuning. Results are averaged over three random seeds.
    \label{tab:ft-breakdown-global}}
\end{table*}

\begin{table*}[tb]
    \centering
    \small
    \setlength\tabcolsep{3.5pt}
    \begin{tabular}{@{}llll@{}}
    \toprule
    \textbf{Model}            & \textbf{Treatment Topics} & \textbf{Control Topics} & \textbf{Unrelated Topics} \\ \midrule
    Before finetuning         & 72.6                     & 91.3     & 61.0                 \\
    Finetuning with baseline & 82.5 (0.9)                & 90.8 (0.3)  & 65.6 (0.8)   \\
    Finetuning with \tool  & \textbf{91.2 (0.5)}         & \textbf{91.2 (0.2)}  & \textbf{74.7 (2.0)}  \\ \bottomrule
    \end{tabular}%
    
    \caption{Accuracies on treatment, control, and unrelated topics. Finetuning on treatment topic improves performance on semantically unrelated bugs within the same set of classes, but gains are smaller than on treatment topics.
    \label{tab:unrelated}}
\end{table*}

\begin{lstlisting}[label=lst:unrelated,caption=Topics from \tool~studied in Sections 4.3 and 4.4 in the main text.,float,frame=htb]
bananagrams
banana in fruit salad
curling broom on the ice
broomball with brooms
candle in mason jar with flower
hexagon candle
a lemon with a smiley face drawn on it
lemon on waffle
lace sandal
crocs
rows of wine bottles in a store
wine bottle in a wine fridge
\end{lstlisting}  

\paragraph{Effect on conceptually unrelated bugs.} We evaluated whether finetuning on treatment topics affected conceptually unrelated bugs. For each class in \sixoverlap, we found two additional topics with high failure rates (Listing \ref{lst:unrelated}), disjoint from the treatment topics in Listing \ref{lst:adatest-topics}. 
We then measured performance on these unrelated topics before and after finetuning, and we compare to performance changes on the treatment and control topics in Table \ref{tab:unrelated}.
We see that finetuning on treatment topic improves performance on semantically unrelated bugs within the same set of classes, but gains are smaller than on treatment topics.




\begin{lstlisting}[label=lst:dominobert-vit,caption=Slice descriptions generated by \dominobert~for target model \vit. Asterisks in front of coherent topics.,float=*,frame=tbh]
a photo of setup by banana (banana)
*a photo of wine and banana (banana)
*a photo of ceramics and banana (banana)
*a photo of basket and banana (banana)
*a photo of munitions and broom (broom)
a photo of activist david broom (broom)
a photo of synthesizer. broom (broom)
*a photo of wildlife at broom (broom)
a photo of violinist jenny broom (broom)
*a photo of literature and candle (candle)
a photo of panchayats candle (candle)
*a photo of altarpiece and candle (candle)
a photo of rob and candle (candle)
a photo of corella lemon (lemon)
*a photo of blender lemon (lemon)
a photo of promoter david lemon (lemon)
a photo of clown billy lemon (lemon)
a photo of estadio jose lemon (lemon)
a photo of rowing on sandal (sandal)
a photo of nana and sandal (sandal)
a photo of placental sandal (sandal)
a photo of skiing at sandal (sandal)
a photo of screenwriter michael sandal (sandal)
a photo of shelter and wine bottle (wine bottle)
*a photo of champange wine bottle (wine bottle)
*a photo of bakery and wine bottle (wine bottle)
*a photo of advertisement on wine bottle (wine bottle)
*a photo of grocery stores wine bottle (wine bottle)
\end{lstlisting}  

\begin{lstlisting}[label=lst:dominobert-resnet,caption=Slice descriptions generated by \dominobert~for target model \resnet. Asterisks in front of coherent topics.,float=*,frame=tbh]
*a photo of ceramics and banana (banana)
a photo of reception by banana (banana)
*a photo of orange and banana (banana)
a photo of neutron star banana (banana)
a photo of architect paul banana (banana)
a photo of singer jenny broom (broom)
a photo of rowing on broom (broom)
*a photo of ornate old broom (broom)
*a photo of factory of broom (broom)
*a photo of primate and broom (broom)
*a photo of blowing the candle (candle)
a photo of consultant john candle (candle)
*a photo of colorful birds candle (candle)
a photo of swamy candle (candle)
*a photo of candle in entryway (candle)
red lemonade series. (lemon)
liz lemon and the observer (lemon)
keith lemon and david bowie (lemon)
liz lemon and the batman (lemon)
a photo of lemon bay shuttle (lemon)
a photo of cognitive development sandal (sandal)
a photo of drilling the sandal (sandal)
a photo of lecture at sandal (sandal)
*a photo of frozen black sandal (sandal)
a photo of longevity by sandal (sandal)
*a photo of golden wine bottle (wine bottle)
a photo of autopsy wine bottle (wine bottle)
*a photo of home and wine bottle (wine bottle)
a photo of libertarian wine bottle (wine bottle)
a photo of estadio wine bottle (wine bottle)
\end{lstlisting}

\begin{lstlisting}[label=lst:dominoofa-vit,caption=Slice descriptions generated by \dominoofa~for target model \vit. Asterisks in front of coherent topics.,float=*,frame=tbh]
a plate on a table with knives and forks (banana)
*a banana next to a bottle of wine and a glass (banana)
*a kitchen counter with bananas and a pineapple on it (banana)
*a banana and two pears in a red basket (banana)
*a large banana sitting in the middle of an abandoned building (banana)
three mops and a bucket against a brick wall (broom)
a man holding a baseball bat in a room (broom)
a woman in a blue dress is looking at a computer (broom)
a green praying mantis standing on a piece of wood (broom)
*a woman sitting on a chair holding a broom (broom)
*a carved pumpkin with a candle in the middle (candle)
*a candle sitting on the ground on a brick floor (candle)
*a group of lit candles in front of a stained glass window (candle)
*a man sitting at a table with candles (candle)
a white bird perched on a tree branch eating (lemon)
*a pitcher pouring lemonade into a glass with lemons (lemon)
three oranges and an apple on a white background (lemon)
a display of tomatoes and other vegetables (lemon)
a bunch of oranges sitting on top of a table (lemon)
a pair of shoes sitting on top of a skateboard (sandal)
a woman holding a small child on her lap (sandal)
*a pink crocheted sandal with a flower on it (sandal)
*a person is wearing a black sandal on their foot (sandal)
a woman laying on a bed with a laptop (sandal)
a group of small bottles of liquor on a table (wine bottle)
*a bottle of champagne in a bowl on a table (wine bottle)
*a bottle of wine and a paper on a counter (wine bottle)
*a glass of wine and a bottle on a table (wine bottle)
*a bottle of wine and a cigar on a table (wine bottle)
\end{lstlisting}      

\begin{lstlisting}[label=lst:dominoofa-resnet,caption=Slice descriptions generated by \dominoofa~for target model \resnet. Asterisks in front of coherent topics.,float=*,frame=tbh]
*a green bowl with some bananas and a piece of fruit (banana)
a plate on a table with knives and forks (banana)
*a bowl of oranges and bananas on a table (banana)
*a banana and two pears in a red basket (banana)
*a large banana sitting in the middle of an abandoned building (banana)
*a broom sitting on the floor in front of a wooden door (broom)
a group of people holding a large wooden stick (broom)
*a close up of a broom with a wooden handle (broom)
three mops and a bucket against a brick wall (broom)
*a broom hanging on the side of a porch (broom)
*a baby girl sitting in front of a birthday cake with a candle (candle)
*a little girl is holding a candle and looking up (candle)
*a group of lit candles in front of a stained glass window (candle)
*a candle sitting on the ground on a brick floor (candle)
*a candle sitting on top of a wooden table (candle)
a group of sliced oranges and kiwi fruit (lemon)
*a pitcher pouring juice into a glass with lemons (lemon)
three oranges and an apple on a white background (lemon)
*a lemon with a smiley face drawn on it (lemon)
*a bowl filled with oranges and a lemon (lemon)
a pair of snoopy shoes and a box on a green table (sandal)
*a woman is wearing a pair of sandals on her feet (sandal)
*a woman wearing sandals standing on a concrete floor (sandal)
a pair of shoes sitting on top of a magazine (sandal)
*two pictures of a woman wearing a pair of sandals (sandal)
*a vase of roses and two bottles of wine (wine bottle)
*a cake with a bottle of wine in a box (wine bottle)
*a bottle of wine and grapes on a counter with a glass (wine bottle)
*a woman next to a row of wine bottles (wine bottle)
*a bottle of wine and a cigar on a table (wine bottle)
\end{lstlisting}


\end{document}